\documentclass[]{dataflow}

\usepackage[toc,page,header]{appendix}

\usepackage{minitoc}
\usepackage{multirow}     
\usepackage{multicol}     
\usepackage{booktabs}     
\usepackage{colortbl}     
\usepackage[table]{xcolor} 
\usepackage{makecell}     
\usepackage{graphicx}     
\usepackage{amssymb}
\usepackage{pifont} 
\usepackage{minted}
\usepackage{fontawesome5}
\title{DataFlex: A Unified Framework for Data-Centric Dynamic Training of Large Language Models}

\author[*, \dagger]{Hao Liang}
\author[*]{Zhengyang Zhao}
\author[*]{Mingrui Chen}
\author[*]{Meiyi Qiang}
\author[]{Lu Ma}
\author[]{Rongyi Yu}
\author[]{Hengyi Feng}
\author[]{Shixuan Sun}
\author[]{Zimo Meng}
\author[]{Xiaochen Ma}
\author[]{Xuanlin Yang}
\author[]{Qifeng Cai}
\author[]{Ruichuan An}
\author[]{Bohan Zeng}
\author[]{Zhen Hao Wong}
\author[]{Chengyu Shen}
\author[]{Runming He}
\author[]{Zhaoyang Han}
\author[]{Yaowei Zheng}
\author[]{Fangcheng Fu}
\author[]{Conghui He}
\author[]{Bin Cui}
\author[]{Zhiyu Li}
\author[]{Weinan E}
\author[\ddagger]{Wentao Zhang}

\affiliation[]{$^{1}$Peking University, $^{2}$University of Chinese Academy of Sciences, $^{3}$Institute for Advanced Algorithms Research, Shanghai, $^{4}$OriginHub Technology, $^{5}$LLaMA-Factory Team, $^{6}$Zhongguancun Academy, $^{7}$OpenDataLab, Shanghai Artificial Intelligence Laboratory, $^{8}$Shanghai Jiao Tong University}

\abstract{

Data-centric training has emerged as a promising direction for improving large language models (LLMs) by optimizing not only model parameters but also the selection, composition, and weighting of training data during optimization. However, existing approaches to data selection, data mixture optimization, and data reweighting are often developed in isolated codebases with inconsistent interfaces, hindering reproducibility, fair comparison, and practical integration.
In this paper, we present \textsc{DataFlex}, a unified data-centric dynamic training framework built upon LLaMA-Factory. \textsc{DataFlex} supports three major paradigms of dynamic data optimization: dynamic sample selection, domain mixture adjustment, and sample reweighting, while remaining fully compatible with the original LLaMA-Factory training workflow. Instead of introducing an external pipeline, \textsc{DataFlex} provides extensible trainer abstractions and modular algorithm components, enabling it to serve as a drop-in replacement for standard LLM training.
Furthermore, \textsc{DataFlex} unifies common model-dependent operations required by data-centric methods, including embedding extraction, model inference, and gradient computation, and supports large-scale training settings such as DeepSpeed ZeRO-3.
We conduct comprehensive experiments covering seven data selection strategies, three data mixture methods, and one data reweighting method. For data selection, dynamic methods consistently outperform static full-data training on MMLU across Mistral-7B and Llama-3.2-3B, with additional evaluation on Qwen3-14B, Llama-3.1-70B, and within-family and cross-benchmark settings. For data mixture, DoReMi and ODM improve both MMLU accuracy and corpus-level perplexity over default data proportions when pretraining Qwen2.5-1.5B on SlimPajama at 6B and 30B token scales, with the same ordering reproduced on Llama-3.2-3B at 6B; we further evaluate MoE-SFT as an online data mixer. Additionally, \textsc{DataFlex} achieves consistent runtime improvements over original implementations.
These results demonstrate that \textsc{DataFlex} provides an effective, efficient, and reproducible infrastructure for studying and deploying data-centric dynamic training methods for LLMs.

}

\date{\today}

\def\emailicon{\raisebox{-1.5pt}{\includegraphics[height=1.05em]{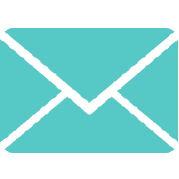}}}
\def\githubicon{\raisebox{-1.5pt}{\includegraphics[height=1.05em]{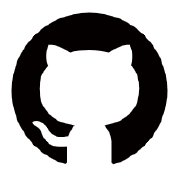}}}
\def\huggingfaceicon{\raisebox{-1.5pt}{\includegraphics[height=1.05em]{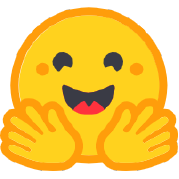}}}

\contribution[*]{Equal Contribution}
\contribution[\dagger]{Project Leader}
\contribution[\ddagger]{Corresponding author}

\checkdata[ \emailicon \hspace{0.3em} Correspondence ]{\email{wentao.zhang@pku.edu.cn}}

\newcommand{\sourcelink}{https://github.com/OpenDCAI/DataFlex}
\newcommand{\datalink}{https://huggingface.co/collections/OpenDCAI/data-for-dataflex}

\checkdata[ \githubicon \hspace{0.3em} Source Code ]{ \url{\sourcelink} }
\checkdata[ \huggingfaceicon \hspace{0.3em} Datasets \& Demo ]{ \url{\datalink} }
\checkdata[ \faFile \hspace{0.57em} Codebase Documentation ]{ \url{https://opendcai.github.io/DataFlex-Doc/} }

\begin{document}
\maketitle

\renewcommand{\thefootnote}{\fnsymbol{footnote}} 
\setcounter{footnote}{0}


\renewcommand{\thefootnote}{\arabic{footnote}}
\pagestyle{fancy}
\fancyhf{}
\fancyhead[L]{DataFlex Technical Report}
\fancyhead[R]{\thepage}

\newpage
\setcounter{tocdepth}{1}
\tableofcontents
\newpage

\section{Introduction}

The remarkable progress of large language models (LLMs)~\cite{llama,qwen} is driven not only by advances in model architectures and optimization algorithms, but also by the scale, quality, and composition of training data. 
As LLM training increasingly relies on massive and heterogeneous corpora, data is no longer merely a static resource, but a critical factor that directly influences training efficiency, generalization, and downstream performance. 
This shift has stimulated increasing interest in data-centric training methods~\cite{xia2024less, wangnice, wang2026opus}, which aim to optimize not only model parameters, but also the selection, composition, and weighting of training data during optimization.

Recent work has explored a diverse set of strategies along this direction. 
Online methods operate within the training loop: data selection approaches such as LESS~\cite{xia2024less} and NICE~\cite{wangnice} estimate sample utility via gradient-based influence or black-box optimization signals, online data mixture methods such as ODM~\cite{albalak2023efficient} adjust domain proportions during pretraining, and data reweighting methods~\cite{sow2025dynamic} modulate per-sample contributions to the training objective. 
Offline methods, by contrast, compute data-centric decisions before training begins: distribution-based selection methods such as TSDS~\cite{liu2024tsds} score and rank samples in an embedding space, while DoReMi~\cite{xie2023doremi} derives optimized domain weights through a proxy model and applies them as a static mixture in the main training run. 
Collectively, these approaches span a wide range of paradigms and design choices, forming a rapidly growing and highly active research landscape around data-centric optimization for LLM training.

However, this diversity also leads to significant fragmentation across methods and implementations.
Most approaches are released as method-specific repositories with inconsistent interfaces, heterogeneous training and evaluation protocols, and often outdated codebases, hindering reproducibility and fair comparison under a unified experimental setting, as summarized in Table~\ref{tab:dataflex_taxonomy} and the appendix. 
As a result, despite the proliferation of algorithms, the community still lacks a unified framework that enables systematic evaluation, reproducible experimentation, and practical deployment of data-centric training methods in modern LLM pipelines.

More fundamentally, existing methods are often designed as isolated algorithmic components rather than as part of a general \emph{data-model interaction} system. 
In practice, many of these methods, whether online or offline, require access to model-dependent signals such as sample embeddings, inference outputs, validation feedback, and gradients. 
Without a unified abstraction to manage such interactions, integrating diverse data-centric training algorithms into scalable workflows remains challenging. 
This limitation becomes even more pronounced in large-scale settings, where efficiency, modularity, and compatibility with existing training infrastructures are critical.

To address these challenges, we introduce the concept of a \textbf{Data-Centric Dynamic Training System} and present \textsc{DataFlex} as its concrete realization~\cite{liang2026towards}. 
A Data-Centric Dynamic Training System is a training infrastructure that treats data as a first-class optimization variable, providing unified interfaces for dynamically controlling which samples are used, how different data sources are mixed, and how strongly each sample contributes to optimization. 
Importantly, the term dynamic here refers to the system's capability to flexibly orchestrate data usage throughout the training lifecycle, rather than a restriction to online-only algorithms: the system accommodates both online methods that adapt data decisions during training and offline methods that precompute data-centric strategies before training begins.

Built upon the widely used LLaMA-Factory~\cite{zheng2024llamafactory} framework, \textsc{DataFlex} serves as a drop-in replacement for the training layer, enabling dynamic control over data usage while preserving compatibility with existing model management, data processing, and optimization pipelines. 
Rather than introducing a disconnected external workflow, \textsc{DataFlex} integrates data-centric strategies directly into the training loop through a modular architecture.

Specifically, \textsc{DataFlex} unifies three representative paradigms of data-centric optimization: data selection, data mixture optimization, and data reweighting. 
To support these paradigms, it provides three corresponding trainer abstractions: Select Trainer, Mix Trainer, and Weight Trainer, along with extensible algorithm components such as selectors, mixers, and weighters.
This design enables diverse methods, regardless of whether they are online or offline, to be implemented under a unified interface, facilitating fair comparison across algorithms and reducing the engineering overhead required to extend the framework.
Moreover, \textsc{DataFlex} standardizes shared model-dependent operations commonly required by data-centric methods, including embedding extraction, model inference, and gradient computation, making it suitable for both research and large-scale training scenarios.

Overall, \textsc{DataFlex} serves two complementary roles. 
First, it provides a reproducible research infrastructure for systematically studying and comparing data-centric training algorithms, both online and offline, within a unified framework. 
Second, it acts as a practical system for improving model training through flexible control over data usage, including which data are selected, how different sources are mixed, and how individual samples influence optimization.

Our main contributions are summarized as follows:

\begin{itemize}
    \item We introduce the concept of a \textbf{Data-Centric Dynamic Training System} and present \textsc{DataFlex}, a unified framework that integrates data selection, data mixture optimization, and data reweighting, covering both online and offline algorithms, into a single training paradigm, enabling systematic comparison and practical deployment.

    \item We propose a \textbf{modular data-model interaction architecture} built upon LLaMA-Factory, with unified trainer abstractions and pluggable algorithm components, enabling flexible and scalable integration of diverse data-centric strategies into standard training workflows.

    \item We develop a \textbf{reproducible and scalable system implementation} that standardizes shared model-dependent operations, including embedding extraction, model inference, and gradient computation, facilitating efficient development and large-scale training.

    \item \textbf{The first evaluation platform for data-centric LLM training}, integrating ten representative algorithms (six data selection, three data mixture, one data reweighting) under a single evaluation protocol with shared seeds, data splits, hardware, and metrics, enabling apples-to-apples comparison that existing fragmented codebases cannot support.
\end{itemize}
\section{Related Work}
\label{sec:related}

\paragraph{Data Selection.}
Selection methods can be \emph{offline}, where the subset is fixed before training, or \emph{online}, where the policy is updated based on the current model state. Offline approaches rely on quality~\citep{du2023mods, chen2023alpagasus}, complexity~\citep{xu2023rethinking, liu2023makes}, or distribution~\citep{liu2024tsds} signals to score samples once. Online approaches (LESS~\citep{xia2024less}, NICE~\citep{wangnice}, LearnAlign~\citep{li2025learnalign}, and loss-based heuristics) adapt to the model's evolving state but require tighter integration with the training loop and efficient access to intermediate signals.

\paragraph{Data Mixture.}
Mixture methods control how multi-domain corpora are sampled. Offline methods such as DoReMi~\citep{xie2023doremi}, DoGE~\citep{fan2023doge}, and RegMix~\citep{liu2024regmix} estimate domain weights via proxy models or bilevel optimization before the main run. Online methods such as ODM~\citep{albalak2023efficient}, Aioli~\citep{chen2024aioli}, and Adaptive Data Optimization~\citep{jiang2024adaptive} adjust mixtures during training based on real-time loss signals. Beyond loss-based signals, MoE-SFT~\citep{zhu2024dynamicdatamixingmaximizes} targets mixture-of-experts models and adjusts domain weights online from gate-load divergence across domains.

\paragraph{Data Reweighting.}
Sample reweighting modulates per-example contributions to the gradient without changing the dataset~\citep{ren2018learning, shu2019meta}. Recent work scales these ideas to LLM training: \citet{thakkar2023self} use self-influence signals, \citet{chen2024hard} emphasize hard samples, \citet{sow2025dynamic} propose dynamic loss-based reweighting, and \citet{adapt2026} unify selection, mixture, and reweighting under a common reweighting formalism, but no system to date integrates all three under a single training framework.

\paragraph{Data-Centric Training Systems.}
Mixtera~\citep{bother2025mixtera} provides a declarative data plane for specifying mixtures and ordering at the data-access layer, but does not provide algorithmic implementations. In contrast, \textsc{DataFlex} integrates pluggable algorithm components directly into the optimization loop on top of LLaMA-Factory~\citep{zheng2024llamafactory}, unifying selection, mixture, and reweighting under a single training framework with reusable trainer abstractions and shared system support for full-rank gradient acquisition under DeepSpeed ZeRO-3. Orthogonally to model-evaluation platforms such as LM-Eval-Harness and HELM, which benchmark models \emph{after} training, \textsc{DataFlex} benchmarks the training-time data interventions that shape those models in the first place. Extended discussion is provided in Appendix~\ref{app:related}.
\begin{figure*}
    \centering
    \includegraphics[width=\linewidth]{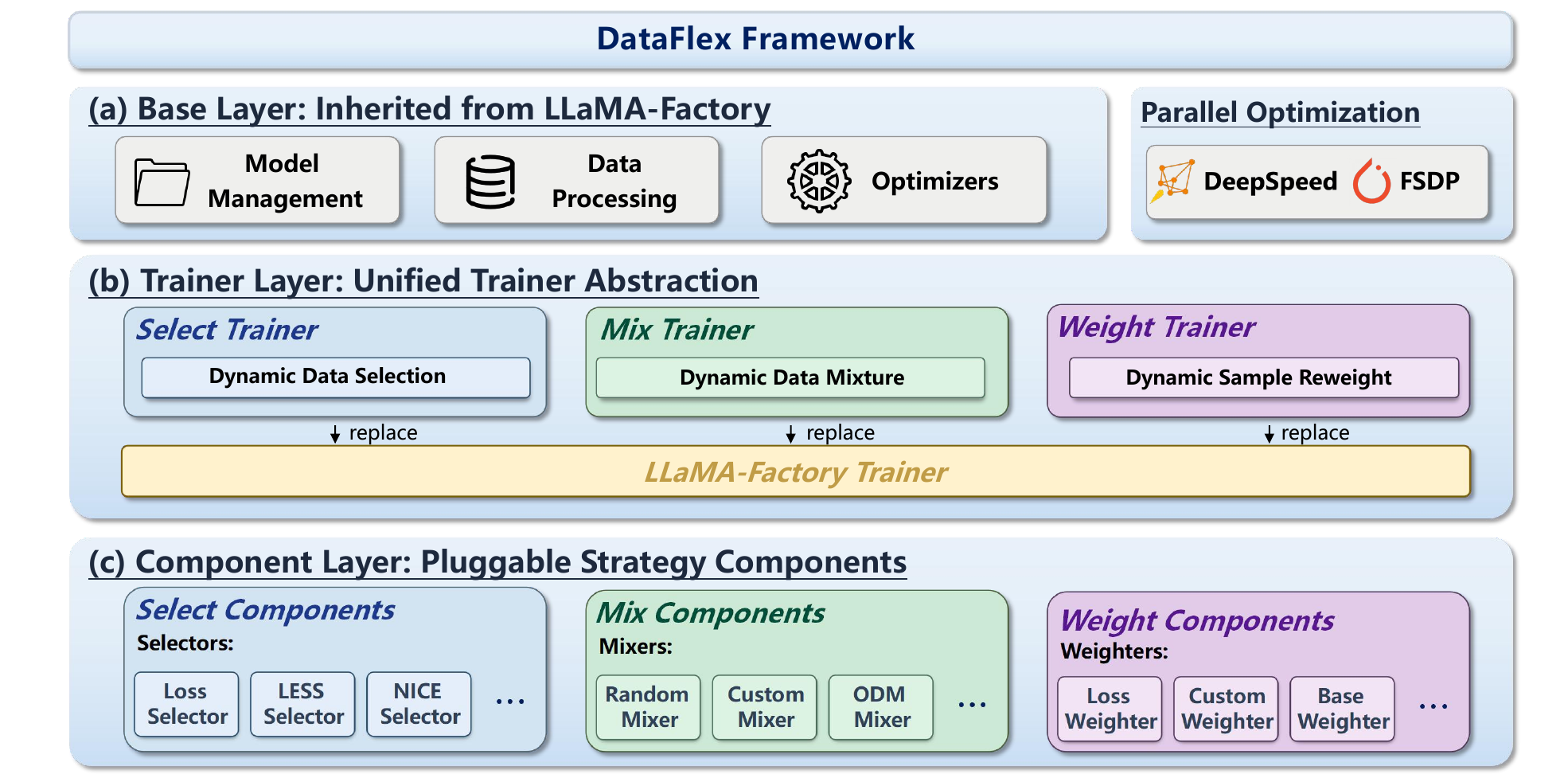}
\caption{
High-level architecture of \textsc{DataFlex}. Each trainer is coupled with a set of pluggable algorithm components, including selectors (e.g., LESS, NICE, loss-based, and delta-loss methods), mixers (e.g., DoReMi, ODM, and MoE-SFT), and weighters (e.g., loss-based weighting).
}
    \label{fig:DataFlex_Overview}
\end{figure*}
\vspace{4mm}
\section{\textsc{DataFlex}: System Design and Core Abstractions}
\label{sec:framework}
\subsection{Goals and Design Philosophy}

\textsc{DataFlex} is a data-centric dynamic training framework for large language models. 
Its central goal is to elevate data from a static training resource to a first-class optimization variable. 
Conventional training pipelines typically rely on fixed datasets, static sampling orders, and predefined domain proportions. 
In contrast, \textsc{DataFlex} enables dynamic control over which samples are selected, how data from different sources are mixed, and how strongly each sample contributes to gradient updates.

The design of \textsc{DataFlex} is guided by three principles. 
First, \textbf{unification}: representative paradigms of data-centric training, including data selection, data mixture optimization, and data reweighting, are supported under a common training framework. 
Second, \textbf{compatibility}: the system integrates seamlessly with existing large-scale model training infrastructure rather than introducing an isolated workflow. 
Third, \textbf{extensibility}: researchers can implement and compare new data-centric algorithms with minimal engineering overhead.

To achieve these goals, \textsc{DataFlex} is built upon LLaMA-Factory and is designed to replace the training layer while preserving the underlying model management and optimization components. 
For data mixture optimization scenarios that require dynamic domain-level resampling, \textsc{DataFlex} further extends the data-loading pipeline with a lightweight adapter. 
This design allows \textsc{DataFlex} to serve as a drop-in replacement for standard training workflows, while providing flexible interfaces for dynamic control over data usage during optimization.

In addition, data-centric training often requires repeated access to intermediate model signals, such as sample embeddings, inference outputs, and gradients. 
\textsc{DataFlex} therefore unifies the management of these operations, enabling reusable implementations of dynamic training algorithms and facilitating deployment in large-scale training settings.

\subsection{Data Module Design}
\label{sec:architecture}
To support dynamic data-centric training, \textsc{DataFlex} introduces a modular architecture consisting of three layers: the base layer, the trainer layer, and the component layer.

\textbf{Base Layer.}
The base layer is inherited from LLaMA-Factory and provides standard infrastructure for model management, data processing, and optimization. 
By reusing this mature backbone, \textsc{DataFlex} avoids reimplementing generic training functionality and remains compatible with existing model fine-tuning workflows.

\textbf{Trainer Layer.}
At the core of \textsc{DataFlex} is a unified trainer abstraction that replaces the original LLaMA-Factory trainer with three dynamic training modes:
(1) \emph{Select Trainer}, which dynamically selects a subset of samples according to a specified strategy;
(2) \emph{Mix Trainer}, which dynamically adjusts mixture ratios across domains or data sources during training; and
(3) \emph{Weight Trainer}, which dynamically modifies per-sample training weights during backpropagation.
These trainers correspond to the three main paradigms of data-centric optimization considered in this work.

\textbf{Component Layer.}
Each trainer is associated with a set of pluggable strategy components.
Specifically, the Select Trainer uses \emph{selectors}, the Mix Trainer uses \emph{mixers}, and the Weight Trainer uses \emph{weighters}. 
These components encapsulate algorithm-specific logic while sharing a common interface within the training framework.
For example, selection components may score or rank samples, mixture components compute domain-level proportions, and weighting components assign scalar coefficients to training instances.
All components are managed through a centralized registry, allowing new algorithms to be registered via decorators and automatically discovered and instantiated by the trainer at runtime.

Overall, the architecture of \textsc{DataFlex} is intentionally lightweight: rather than introducing an external orchestration layer on top of LLaMA-Factory, it replaces the training layer and introduces only minimal extensions elsewhere (e.g., the data-loading pipeline for mixture scenarios). 
This design preserves compatibility while enabling a unified implementation of dynamic data-centric training.

\begin{figure*}
    \centering
    \includegraphics[width=0.9\linewidth]{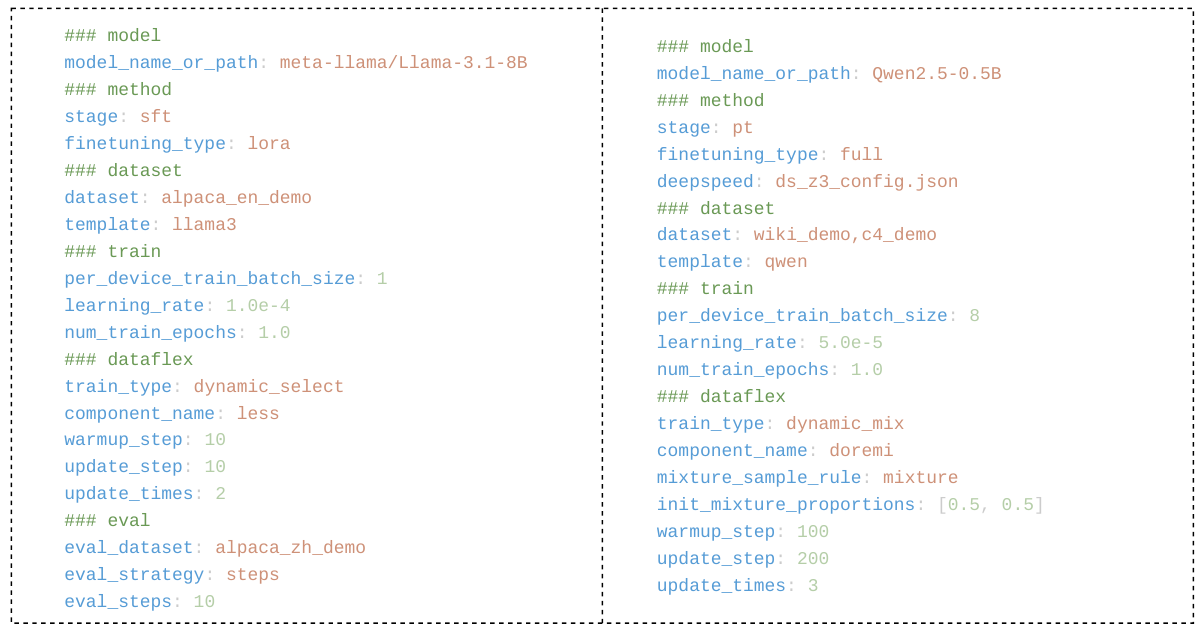}
\caption{
Example configurations in \textsc{DataFlex}: dynamic data selection using LESS (left) and dynamic data mixture using DoReMi (right). Both configurations share the same LLaMA-Factory--compatible structure for model, dataset, and training settings, differing only in the \texttt{dataflex} section, which demonstrates the drop-in design of the framework.
}
    \label{fig:config_examples}
\end{figure*}
\subsection{Algorithm Integration and Extensibility}
\label{sec:algorithms}
\paragraph{Unified Trainer--Component Interaction.}
Although data selection, data mixture, and data reweighting differ in their control targets, they share a common interaction pattern: each method observes the current model state, computes a data-centric decision, and feeds it back into subsequent optimization steps.
\textsc{DataFlex} abstracts this pattern into a standardized trainer--component interaction.
During training, each trainer invokes its corresponding component to produce a control signal: for data selection, the signal is a subset or ranking of candidate samples; for data mixture, it is a set of domain-level mixing ratios; and for data reweighting, it is a set of sample-wise weights.
The invocation frequency is tailored to each paradigm: the Select Trainer and Mix Trainer invoke their components at configurable intervals (controlled by \texttt{warmup\_step} and \texttt{update\_step}), while the Weight Trainer applies per-sample weighting at every training step after a warmup phase.
This unified abstraction decouples algorithm-specific logic from the training pipeline. A researcher can implement a new selector, mixer, or weighter as a self-contained component and register it via the registry without modifying the rest of the system.

\paragraph{Supported Algorithms.}
Table~\ref{tab:dataflex_taxonomy} summarizes the data-centric methods currently integrated into \textsc{DataFlex}.
For data selection, the framework supports gradient-based methods (LESS~\cite{xia2024less}, NICE~\cite{wangnice}), loss-based methods (Loss, Delta Loss~\cite{zhang2025preference}), and distribution-based methods (NEAR, TSDS~\cite{liu2024tsds}).
For data mixture, it includes offline DoReMi~\cite{xie2023doremi} and online ODM~\cite{albalak2023efficient} loss-based methods, as well as MoE-SFT~\cite{zhu2024dynamicdatamixingmaximizes}, an online method that operates on a distinct gate-load signal specific to mixture-of-experts models rather than on loss. The inclusion of MoE-SFT demonstrates that the mixer interface generalizes beyond loss-based feedback to architecture-specific signals.
For data reweighting, a loss-based weighter with multiple strategies is provided.
All methods are implemented as pluggable components under a unified interface, enabling fair comparison under controlled conditions.

\paragraph{Configuration and Usability.}
A key design goal of \textsc{DataFlex} is to minimize the effort required to switch from standard training to dynamic data-centric training.
As illustrated in Figure~\ref{fig:config_examples}, a \textsc{DataFlex} configuration file reuses the same YAML-based format as LLaMA-Factory for specifying the model, dataset, and training hyperparameters.
The only addition is a short \texttt{dataflex} section that tells the framework \emph{what kind} of data-centric strategy to apply and \emph{how} to schedule it.
Concretely, users specify three fields:
(1)~\texttt{train\_type}, which chooses the training paradigm---for example, \texttt{dynamic\_select} for data selection or \texttt{dynamic\_mix} for data mixture;
(2)~\texttt{component\_name}, which picks the specific algorithm to use, such as \texttt{less} or \texttt{doremi}; and
(3)~a small set of scheduling parameters (\texttt{warmup\_step}, \texttt{update\_step}, \texttt{update\_times}) that control when and how often the data-centric strategy is updated during training.
Paradigm-specific options can also be added where needed; for instance, data mixture configurations accept \texttt{init\_mixture\_proportions} to set the initial domain weights.

In practice, the workflow is nearly identical to standard LLaMA-Factory usage.
After installing \textsc{DataFlex} alongside LLaMA-Factory via \texttt{pip install -e .}, users launch training with \texttt{dataflex-cli train <config.yaml>}, which mirrors the familiar \texttt{llamafactory-cli train <config.yaml>} command.
In other words, converting an existing LLaMA-Factory configuration to a data-centric one only requires appending the \texttt{dataflex} section---no changes to the model, data, or optimizer settings are needed.

\subsection{System Efficiency and Scalability}
\label{sec:system_support}
By building upon LLaMA-Factory, \textsc{DataFlex} inherits support for mixed-precision training, distributed data parallelism, and DeepSpeed integration, introducing minimal overhead while enabling data-centric control.

A key challenge lies in gradient acquisition: many data selection and reweighting methods require full parameter gradients, which are not typically materialized under model parallelism.
\textsc{DataFlex} addresses this challenge with a distributed gradient collection mechanism compatible with DeepSpeed ZeRO-3, leveraging the \texttt{safe\_get\_full\_grad} interface to reconstruct full gradients from partitioned shards, and \texttt{safe\_get\_full\_optimizer\_state} to access optimizer states when needed.

To reduce overhead, \textsc{DataFlex} executes gradient-based operations at configurable intervals rather than at every step, caches selection and weighting decisions for reuse across multiple steps, and supports lightweight proxy signals (e.g., loss values) when full gradients are unnecessary.
All data-centric components operate within the distributed training loop and can be parallelized across workers without centralized coordination, ensuring scalability to multi-node, multi-GPU settings.

\begin{table}[t]
\centering
\small
\caption{Data-centric training methods unified in \textsc{DataFlex}. Symbols: \ding{52} (Yes), \ding{56} (No), $\triangle$ (Partial: an official implementation exists but exhibits usability or stability limitations).}
\label{tab:dataflex_taxonomy}
\setlength{\tabcolsep}{3pt}
\renewcommand{\arraystretch}{1.05}
\resizebox{\linewidth}{!}{
\begin{tabular}{lcccccccccc}
\toprule
\multirow{2}{*}{\textbf{Attribute}}
& \multicolumn{6}{c}{\textbf{Data Selection}}
& \multicolumn{3}{c}{\textbf{Data Mixture}}
& \multicolumn{1}{c}{\textbf{Reweighting}} \\
\cmidrule(lr){2-7} \cmidrule(lr){8-10} \cmidrule(lr){11-11}
& \textbf{LESS} & \textbf{NICE} & \textbf{Loss} & \textbf{Delta Loss} & \textbf{NEAR} & \textbf{TSDS}
& \textbf{DoReMi} & \textbf{ODM} & \textbf{MoE-SFT}
& \textbf{Loss Reweight} \\
\midrule
\textbf{Signal Type}      
& Gradient & Gradient & Loss & Loss & Distribution & Distribution 
& Loss & Loss & Gate Load 
& Loss \\

\textbf{Optimization Mode}
& Online & Online & Online & Online & Offline & Offline 
& Offline & Online & Online 
& Online \\

\textbf{Model-in-the-Loop}
& \ding{52} & \ding{52} & \ding{52} & \ding{52} & \ding{56} & \ding{56} 
& \ding{52} & \ding{52} & \ding{52} 
& \ding{52} \\

\textbf{Official Repo}
& $\triangle$ & $\triangle$ & \ding{56} & \ding{56} & \ding{56} & \ding{52} 
& $\triangle$ & $\triangle$ & \ding{52} 
& \ding{56} \\

\bottomrule
\end{tabular}
}
\end{table}

\section{Experiments}
\label{sec:exp}

We report the three benchmark axes enabled by \textsc{DataFlex}: (i) data selection and reweighting on instruction tuning across multiple backbones and scales, (ii) data mixture on pretraining across two corpus scales and two backbones, and (iii) system efficiency of \textsc{DataFlex}'s reference implementations relative to the original codebases. Every result follows the standardized protocol of Section~\ref{sec:framework}: baselines and proposed methods share the same trainer infrastructure, dataloaders, seeds, and evaluation cadence, so that reported differences reflect the data-centric strategy rather than implementation artifacts. We summarize the most important configuration choices below; complete hyperparameters, hardware specifications, and reproducibility details are provided in Appendix~\ref{app:hparams}.

\subsection{Data Selection and Reweighting}
\label{sec:exp_selection}

\paragraph{Setup.}
We instruction-tune Mistral-7B-v0.1~\citep{2023arXiv231006825J} and Llama-3.2-3B~\citep{grattafiori2024llama} on a 100k-example subset of Open-Hermes-2.5 using LoRA ($r{=}32$, $\alpha{=}64$, all linear layers) with a global batch size of 8. We compare four online selection methods (LESS extended to online, NICE, Loss, Delta Loss), one reweighting method~\citep{sow2025dynamic}, two offline selection methods (NEAR, TSDS), a static full-data baseline, and a Random-selection baseline. Evaluation uses MMLU, with online methods configured with \texttt{warmup\_step}=100, \texttt{update\_step}=50, and \texttt{update\_times}=30. All experiments run on 8$\times$NVIDIA H20 GPUs.

\begin{figure}[t]
  \centering
  \begin{subfigure}[t]{0.49\linewidth}
    \includegraphics[width=\linewidth]{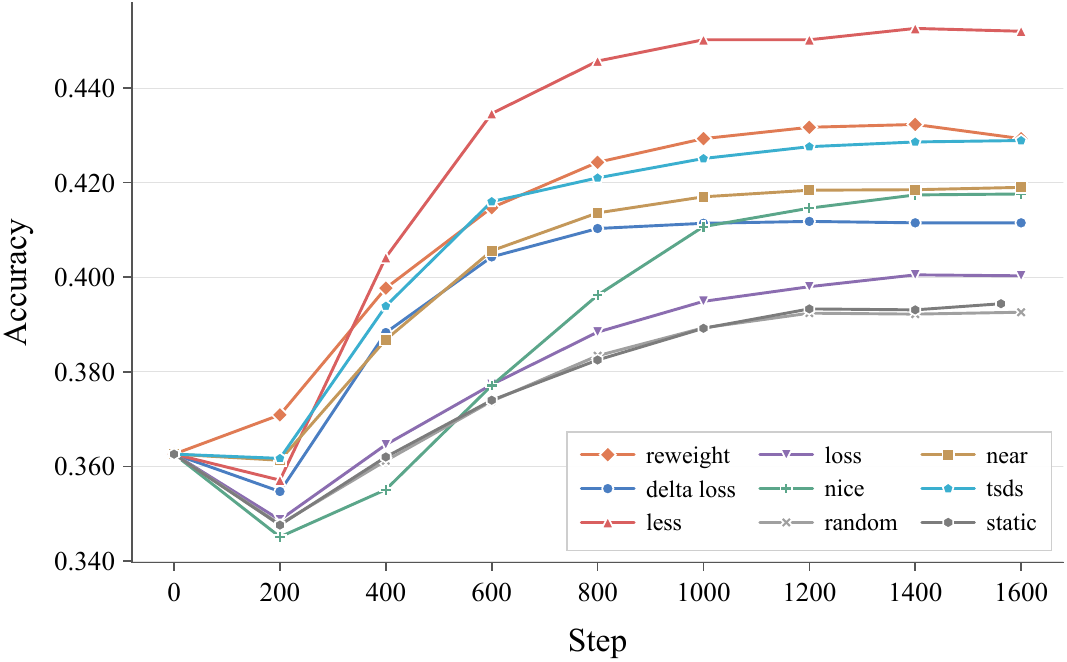}
    \caption{Mistral-7B}
    \label{fig:mistral-7b}
  \end{subfigure}\hfill
  \begin{subfigure}[t]{0.49\linewidth}
    \includegraphics[width=\linewidth]{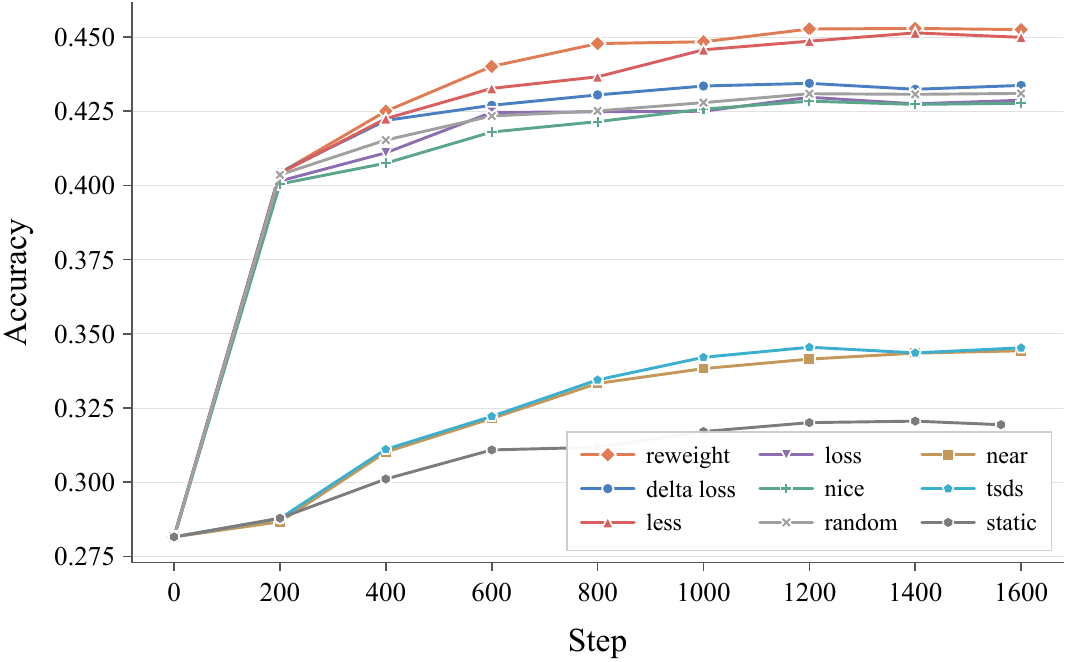}
    \caption{Llama-3.2-3B}
    \label{fig:llama3}
  \end{subfigure}
  \caption{MMLU accuracy under different data selection and reweighting strategies, evaluated every 200 training steps. Dynamic methods that leverage model-dependent signals (LESS and Reweight, joined by Delta Loss on Llama-3.2-3B) consistently outperform the static full-data baseline on both backbones.}
  \label{fig:two-models}
\end{figure}

\paragraph{Results.}
Figure~\ref{fig:two-models} reports MMLU accuracy curves. On \textbf{Mistral-7B}, LESS achieves the best final accuracy (0.452), exceeding the static baseline (0.394) by 5.8 points. Reweight and TSDS tie at 0.429, followed by NEAR (0.419), NICE (0.418), and Delta Loss (0.412). On the smaller \textbf{Llama-3.2-3B}, the gap widens substantially: the static baseline plateaus at 0.319, while Reweight (0.453), LESS (0.450), and Delta Loss (0.434) all exceed the baseline by more than 11 points. Offline methods (NEAR 0.344, TSDS 0.345) perform notably worse on the smaller model than online methods.

\paragraph{Evaluation on larger LLM backbones.}
We further evaluate representative methods on Qwen3-14B (Table~\ref{tab:selection_qwen14b}) and Llama-3.1-70B (Table~\ref{tab:selection_llama70b}). 
On Qwen3-14B, Delta Loss, Reweight, and TSDS outperform Random on all four evaluated benchmarks. 
TSDS improves MATH-500 by 5.60 points over Random. On Llama-3.1-70B, LESS improves over the base model on all six reported benchmarks. 
Delta Loss, Reweight, and TSDS also improve MMLU to 72.55\%, 72.42\%, and 72.54\%, respectively, versus 70.15\% for the base model.

\begin{table}[t]
\centering
\small
\caption{Data selection and reweighting on Qwen3-14B. The base model is evaluated at step 0; other methods are evaluated at step 200. Scores are accuracy (\%).}
\label{tab:selection_qwen14b}
\setlength{\tabcolsep}{5pt}
\renewcommand{\arraystretch}{1.04}
\begin{tabular}{lcccc}
\toprule
\textbf{Method} & \textbf{MMLU} & \textbf{ARC-C} & \textbf{ARC-E} & \textbf{MATH-500} \\
\midrule
Base       & 68.31 & 60.75 & 82.70 & 58.80 \\
Random     & 69.54 & 62.20 & 83.75 & 58.80 \\
Delta Loss & 71.02 & 64.59 & \textbf{85.73} & 60.80 \\
Reweight   & \textbf{72.24} & \textbf{65.19} & 84.72 & 59.60 \\
TSDS       & 72.14 & 65.02 & 84.39 & \textbf{64.40} \\
\bottomrule
\end{tabular}
\end{table}

\begin{table}[t]
\centering
\small
\caption{Cross-benchmark evaluation on Llama-3.1-70B. The base model is evaluated at step 0 and LESS at step 400. Scores are accuracy (\%).}
\label{tab:selection_llama70b}
\setlength{\tabcolsep}{3.5pt}
\renewcommand{\arraystretch}{1.04}
\begin{tabular}{lcccccc}
\toprule
\textbf{Method} & \textbf{MMLU} & \textbf{ARC-C} & \textbf{ARC-E} & \textbf{GSM8K} & \textbf{IFEval} & \textbf{BBH} \\
\midrule
Base & 70.15 & 64.33 & 86.45 & 13.50 & 11.28 & 82.12 \\
LESS & \textbf{72.72} & \textbf{69.54} & \textbf{88.76} & \textbf{18.50} & \textbf{23.29} & \textbf{84.46} \\
\bottomrule
\end{tabular}
\end{table}

\paragraph{Benchmark findings.}
Three patterns emerge from the head-to-head comparison. \textbf{(F1)} Online methods that leverage model-dependent signals (LESS and Reweight, joined by Delta Loss on Llama-3.2-3B) consistently rank at the top on both backbones, suggesting that exploiting the evolving model state is a particularly effective curation strategy in these settings. \textbf{(F2)} The benefit of dynamic selection is \emph{backbone dependent}: the best-to-baseline gap is 5.8 points on Mistral-7B (LESS) and 13.4 points on Llama-3.2-3B (Reweight). Since the two backbones differ in both size and model family, this is a cross-model observation rather than evidence of a capacity-only effect; a within-family comparison on Llama-3.2-1B and Llama-3.2-3B (Appendix~\ref{app:selection_additional}) further shows that the best method varies with scale and benchmark. \textbf{(F3)} Offline methods (NEAR, TSDS) provide a lower-overhead alternative but cannot adapt to the evolving model state, which is especially costly on the smaller Llama-3.2-3B where they trail every online method by roughly 8 points or more. These three findings are directly enabled by the unified protocol and cannot be read off any individual algorithm's original paper. Per-method numerical breakdowns are provided in Appendix~\ref{app:discussion}.

\subsection{Data Mixture}
\label{sec:exp_mixture}

\paragraph{Setup.}
We pretrain Qwen2.5-1.5B~\citep{qwen2.5} from scratch on SlimPajama~\citep{cerebras2023slimpajama} at two scales: SlimPajama-6B and SlimPajama-30B, both sampled to preserve the natural domain proportions (CC 54.1\%, C4 28.7\%, GitHub 4.2\%, Book 3.7\%, ArXiv 3.4\%, Wikipedia 3.1\%, StackExchange 2.8\%). We compare three mixture strategies: a \emph{static baseline} using the natural proportions, \emph{DoReMi}~\citep{xie2023doremi} (offline three-step pipeline using a Qwen2.5-0.5B proxy), and \emph{ODM}~\citep{albalak2023efficient} (online Exp3-based bandit). Evaluation reports MMLU accuracy~\citep{hendryckstest2021} (5-shot) and per-domain log perplexity. The 6B experiments run on 8 H20 GPUs; the 30B experiments scale to 32 H20 GPUs (4 nodes) coordinated via \texttt{torchrun}, all under DeepSpeed ZeRO-3 with BF16.

\begin{table}[t]
\centering
\caption{Comparison of data mixing strategies on SlimPajama. MMLU accuracy (Acc, $\uparrow$) and per-domain log perplexity (PPL, $\downarrow$). Both DoReMi and ODM outperform the static baseline on MMLU; the two methods exhibit complementary strengths across domains and metrics.}
\label{tab:data_mixture}
\setlength{\tabcolsep}{4pt}
\renewcommand{\arraystretch}{0.95}
\small
\begin{tabular}{lcc|ccccccc}
\toprule
\multirow{2}{*}{\textbf{Method}}
& \multicolumn{1}{c}{\textbf{Acc} $\uparrow$}
& \multicolumn{8}{c}{\textbf{Log Perplexity (PPL)} $\downarrow$} \\
\cmidrule(lr){2-2} \cmidrule(lr){3-10}
& \textbf{MMLU}
& \textbf{ALL} & \textbf{CC} & \textbf{C4} & \textbf{SE}
& \textbf{Wiki} & \textbf{GitHub} & \textbf{ArXiv} & \textbf{Book} \\
\midrule
\multicolumn{10}{c}{\textit{\textbf{SlimPajama-6B}}} \\
\midrule
Baseline & 25.27 & 4.217 & 4.278 & 4.532 & 3.402 & \textbf{3.546} & \textbf{2.640} & 3.508 & 4.778 \\
DoReMi   & 25.84 & \textbf{4.134} & \textbf{4.108} & \textbf{4.358} & 3.788 & 3.997 & 3.420 & 3.413 & 4.661 \\
ODM      & \textbf{26.04} & 4.244 & 4.326 & 4.555 & \textbf{3.243} & 3.699 & 2.704 & \textbf{2.904} & \textbf{4.613} \\
\midrule
\multicolumn{10}{c}{\textit{\textbf{SlimPajama-30B}}} \\
\midrule
Baseline & 25.51 & 3.584 & 3.723 & 3.505 & 2.850 & 3.215 & 3.163 & 4.540 & 5.329 \\
DoReMi   & \textbf{25.97} & 3.562 & 3.731 & \textbf{3.503} & 2.706 & 2.985 & 2.973 & 4.441 & 5.214 \\
ODM      & 25.63 & \textbf{3.429} & \textbf{3.598} & 3.519 & \textbf{2.382} & \textbf{2.713} & \textbf{2.255} & \textbf{3.487} & \textbf{4.746} \\
\bottomrule
\end{tabular}
\end{table}

\paragraph{Results.}
Both dynamic methods outperform the static baseline on MMLU at both scales (Table~\ref{tab:data_mixture}). On \textbf{SlimPajama-6B}, ODM achieves the highest MMLU accuracy (26.04\%) and DoReMi achieves the lowest overall log perplexity (4.134); the two methods exhibit \emph{complementary} strengths across domains: DoReMi is best on the two largest domains (CC, C4), while ODM is best on smaller specialized domains (StackExchange, ArXiv, Book). On \textbf{SlimPajama-30B}, both methods also improve overall log perplexity: DoReMi achieves the highest MMLU (25.97\%) and ODM achieves the lowest overall log perplexity (3.429), winning 6 of 7 domains. At the 30B scale, the static baseline no longer wins any domain, indicating that dynamic mixing remains advantageous at the larger token budget.

\paragraph{Evaluation on another backbone.}
To test whether these observations are specific to the Qwen backbone, we repeat the SlimPajama-6B experiment with Llama-3.2-3B~\citep{grattafiori2024llama} as the target model, using Llama-3.2-1B as DoReMi's reference and proxy models (Steps~1--2). As shown in Table~\ref{tab:data_mixture_llama}, DoReMi and ODM improve MMLU over the static baseline by 0.45 and 0.79 points, respectively, and DoReMi again achieves the lowest overall log perplexity (3.358). Both the MMLU ordering (ODM $>$ DoReMi $>$ baseline) and the log-perplexity ordering (DoReMi $<$ baseline $<$ ODM) match those observed for Qwen2.5-1.5B under the same 6B-token budget, providing additional evidence that the mixture findings are not confined to a single model family.

\begin{table}[t]
\centering
\begin{minipage}[c]{0.5\linewidth}
\setlength{\abovecaptionskip}{0pt}
\setlength{\belowcaptionskip}{0pt}
\caption{Data mixing on SlimPajama-6B with Llama-3.2-3B as the target model. DoReMi uses Llama-3.2-1B for its reference and proxy models. MMLU accuracy (\%, $\uparrow$) and overall log perplexity ($\downarrow$). The orderings match those observed for Qwen2.5-1.5B at the same token budget (Table~\ref{tab:data_mixture}).}
\label{tab:data_mixture_llama}
\end{minipage}\hfill
\begin{minipage}[c]{0.5\linewidth}
\centering
\small
\setlength{\tabcolsep}{6pt}
\renewcommand{\arraystretch}{1.04}
\begin{tabular}{lcc}
\toprule
\textbf{Method} & \textbf{MMLU} $\uparrow$ & \textbf{Log-PPL} $\downarrow$ \\
\midrule
Baseline & 25.17 & 3.409 \\
DoReMi   & 25.62 & \textbf{3.358} \\
ODM      & \textbf{25.96} & 3.471 \\
\bottomrule
\end{tabular}
\end{minipage}
\end{table}

\paragraph{Benchmark findings.}
The two methods reflect distinct mechanisms: DoReMi's minimax excess-loss optimization favors high-resource domains, while ODM's online bandit exploration upweights minority domains such as StackExchange, ArXiv, and Book. At 30B, this complementarity shifts from domains to metrics: ODM, with more steps to converge, achieves the best log perplexity on most domains, while DoReMi's optimized weights yield the highest MMLU. Crucially, neither original codebase is reproducible at the scale we report; the unified protocol in \textsc{DataFlex} is what makes this side-by-side observation possible.

\subsection{System Efficiency}
\label{sec:exp_efficiency}

We benchmark \textsc{DataFlex}'s implementations against the original codebases for both the \emph{Data Selection} module (LESS, online gradient-based; TSDS, offline distribution-based) and the \emph{Data Mixture} module (MoE-SFT~\citep{zhu2024dynamicdatamixingmaximizes}, the third mixture method integrated in \textsc{DataFlex}, which we adopt as the reference for this module because the original DoReMi and ODM codebases are not directly reproducible at our scale; see Appendix~\ref{app:moe_sft}). Data Reweighting is omitted because no widely adopted reference implementation exists.

\subsubsection{Data Selection Efficiency}
\label{sec:exp_selection_efficiency}

\paragraph{LESS (online, gradient-based).}
We compare \textsc{DataFlex}'s implementation of LESS against the original codebase on a 100k subset of Open-Hermes-2.5 with \texttt{Llama-2-7b-hf}~\citep{touvron2023llama}. Since the original LESS only supports single-GPU, we restrict both implementations to a single H20 GPU and additionally report \textsc{DataFlex}'s scaling to 8$\times$H20.

\begin{table}[t]
\centering
\caption{Efficiency and accuracy comparison between \textsc{DataFlex} and the original LESS implementation on the MMLU test set. Time reduction is computed relative to the LESS baseline at the same sampling ratio. For the 8-GPU row, the reduction is computed relative to the original LESS at ratio 1.0.}
\label{tab:efficiency_less}
\small
\renewcommand{\arraystretch}{1.08}
\begin{tabular}{l l c c c}
\toprule
\textbf{Ratio} & \textbf{Method} & \textbf{Acc.\ (\%)} & \textbf{Time (s)} & \textbf{Reduction $\downarrow$} \\
\midrule
\multirow{2}{*}{0.05}
& LESS & 34.91 & 1{,}640 & \multirow{2}{*}{\textbf{3.72\%}} \\
& \textsc{DataFlex} & \textbf{38.35} & \textbf{1{,}579} & \\
\midrule
\multirow{2}{*}{0.1}
& LESS & 37.97 & 3{,}735 & \multirow{2}{*}{\textbf{4.34\%}} \\
& \textsc{DataFlex} & \textbf{40.25} & \textbf{3{,}573} & \\
\midrule
\multirow{2}{*}{0.5}
& LESS & \textbf{41.57} & 14{,}398 & \multirow{2}{*}{\textbf{7.09\%}} \\
& \textsc{DataFlex} & 40.93 & \textbf{13{,}377} & \\
\midrule
\multirow{2}{*}{1.0}
& LESS & 40.38 & 30{,}239 & \multirow{2}{*}{\textbf{4.98\%}} \\
& \textsc{DataFlex} & \textbf{42.37} & \textbf{28{,}734} & \\
\midrule
\rowcolor{gray!10}
1.0 & \textsc{DataFlex} (8 GPUs) & \textbf{43.01} & \textbf{12{,}965} & \textbf{57.13\%} \\
\bottomrule
\end{tabular}
\end{table}

\noindent Table~\ref{tab:efficiency_less} shows that \textsc{DataFlex} consistently reduces single-GPU runtime while achieving comparable or higher accuracy: at ratio 0.05, runtime drops by 3.72\% (1{,}640 $\to$ 1{,}579~s) with accuracy improving from 34.91\% to 38.35\%; at ratio 0.5, runtime drops by 7.09\%, saving over 1{,}000 seconds in a single selection--training cycle. Scaling to 8$\times$H20 GPUs at ratio 1.0 reduces runtime from 30{,}239~s (original LESS, single-GPU) to 12{,}965~s, a 57.13\% reduction, while reaching the highest accuracy (43.01\%). The single-GPU speedups arise from cached gradient projections and overlapped dataloader rebuilds, while the 8-GPU reduction stems from distributed gradient collection under DeepSpeed ZeRO-3 (Section~\ref{sec:system_support}), a capability absent in the original implementation.

\paragraph{TSDS (offline, distribution-based).}
We compare \textsc{DataFlex}'s implementation of TSDS against the original operator under two scaling regimes that are characteristic of practical deployment: varying the size of the training pool and varying the size of the validation anchor set. Both implementations exhibit linear growth in both regimes; \textsc{DataFlex}'s implementation is consistently faster.

\begin{figure}[t]
  \centering
  \begin{subfigure}[t]{0.49\linewidth}
    \centering
    \includegraphics[width=\linewidth]{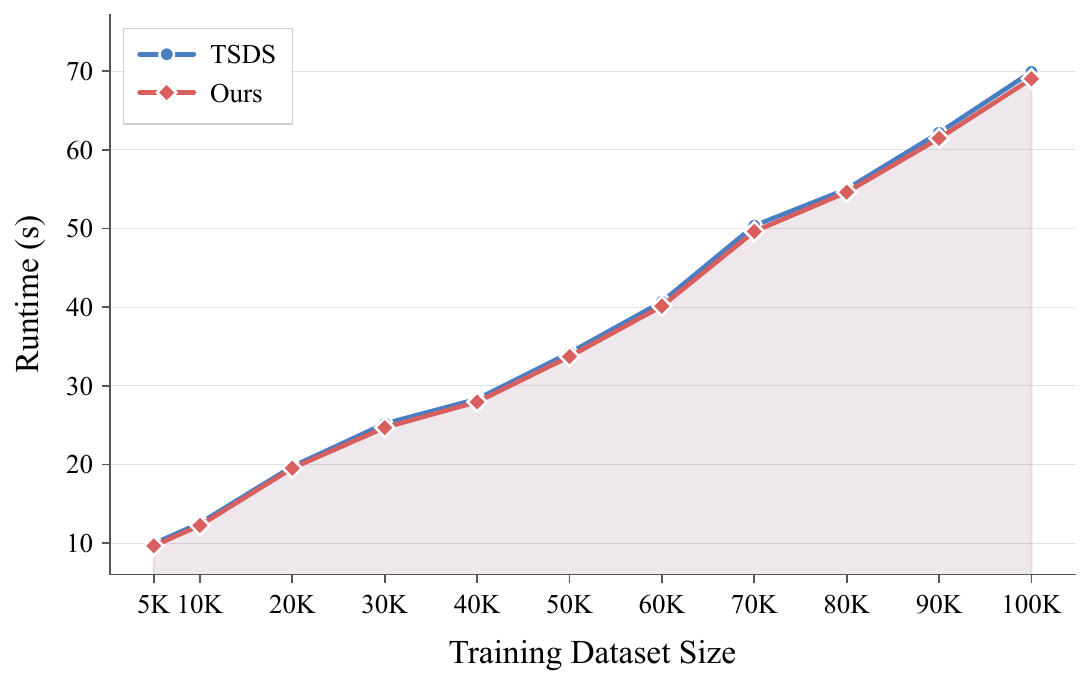}
  \end{subfigure}\hfill
  \begin{subfigure}[t]{0.49\linewidth}
    \centering
    \includegraphics[width=\linewidth]{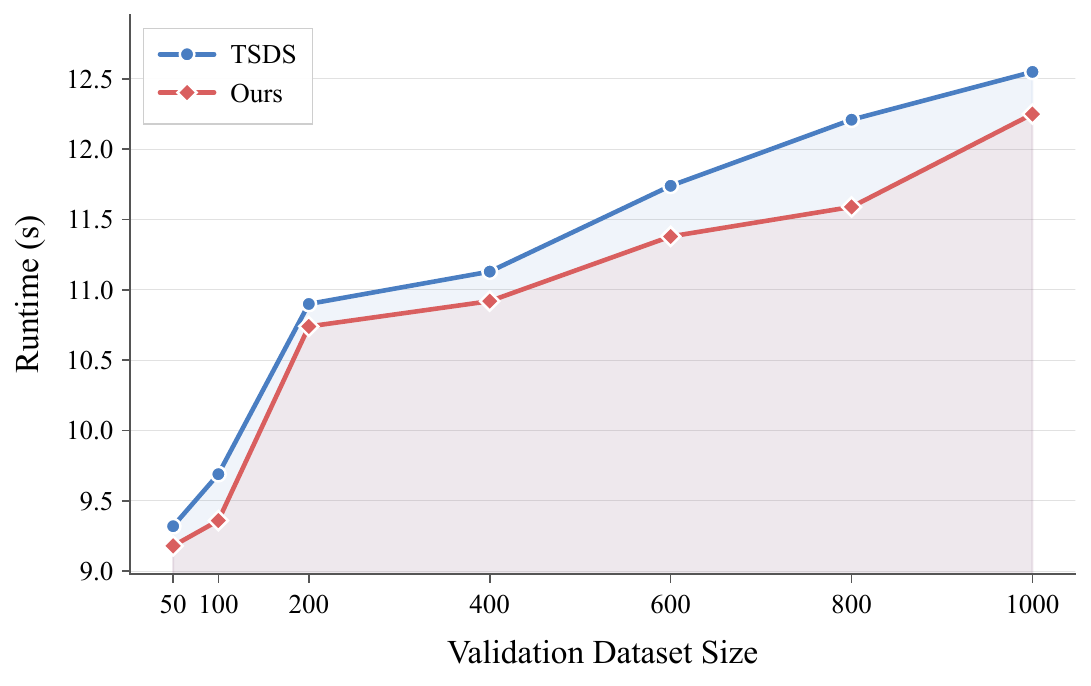}
  \end{subfigure}
  \caption{Efficiency comparison between the original TSDS operator and our implementation under different dataset scaling strategies. Left: the validation set size is fixed while the training set size varies. Right: the training set size is fixed while the validation set size varies.}
  \label{fig:tsds_efficiency}
\end{figure}

\noindent As shown in Figure~\ref{fig:tsds_efficiency}, scaling the training pool from 5k to 100k examples (validation fixed at 1k) reduces runtime from 9.94\,s to 9.65\,s at the smallest scale and from 69.91\,s to 69.03\,s at the largest scale, a steady 1--3\% improvement. Scaling the validation set from 50 to 1{,}000 (training fixed at 10k) yields per-point reductions ranging from about 1.5\% to 5\%, with the peak at 800 queries (12.21\,s $\to$ 11.59\,s). The efficiency advantage is modest at any single scale but consistent, making \textsc{DataFlex} preferable for iterative experimental workflows where TSDS is repeatedly invoked under different configurations.

\subsubsection{Data Mixture Efficiency}
\label{sec:exp_mixture_efficiency}

We additionally evaluate \textsc{DataFlex}'s online data mixer on supervised fine-tuning of mixture-of-experts (MoE) language models, following the MoE-SFT setting~\citep{zhu2024dynamicdatamixingmaximizes}, where the mixer adjusts domain weights using \emph{gate-load} divergence signals (an MoE-specific feedback distinct from the loss signals used by DoReMi and ODM) at every $m$-th step. We train LLaMA-MoE-v1-3.5B-2/8~\citep{zhu2024llamamoebuildingmixtureofexpertsllama} for 2{,}000 steps on four instruction-tuning domains (code, math, ShareGPT, Orca) under DeepSpeed ZeRO-3 with BF16; full algorithmic details and hyperparameters are provided in Appendix~\ref{app:moe_sft}.

\begin{table}[t]
\centering
\caption{Efficiency and effectiveness comparison between the official MoE-SFT~\citep{zhu2024dynamicdatamixingmaximizes} online data mixer and \textsc{DataFlex}'s implementation. Training time ($\downarrow$) reflects efficiency; downstream task scores ($\uparrow$) reflect effectiveness.}
\label{tab:online_data_mixer_comparison}
\setlength{\tabcolsep}{6pt}
\renewcommand{\arraystretch}{1.05}
\begin{tabular}{lcccccc}
\toprule
\textbf{Method} & \textbf{Training Time} $\downarrow$ & \textbf{MMLU} & \textbf{BBH} & \textbf{QA} & \textbf{MBPP} & \textbf{Average} \\
\midrule
Official            & 5:15:19 (18{,}919\,s)         & 32.46          & 30.07          & \textbf{61.74} & 15.09          & 34.84 \\
\textsc{DataFlex}   & \textbf{5:01:02 (18{,}062\,s)} & \textbf{33.16} & \textbf{30.24} & 61.11          & \textbf{16.08} & \textbf{35.15} \\
\bottomrule
\end{tabular}
\end{table}

\noindent Table~\ref{tab:online_data_mixer_comparison} shows that \textsc{DataFlex} reduces wall-clock training time by 4.5\% (18{,}919\,s $\to$ 18{,}062\,s) while matching the official implementation in model quality, with a slightly higher average score (35.15 vs.\ 34.84). Together with the LESS and TSDS results, this indicates that \textsc{DataFlex}'s system-level efficiency benefits generalize across paradigms while preserving underlying mixing dynamics.

\paragraph{Cross-paradigm takeaways.}
Running all three paradigms under a single protocol surfaces regularities that are invisible from any individual algorithm's original paper. \textbf{(T1) Dynamic $>$ static holds across the evaluated paradigms:} online selection (F1), offline/online mixture (§\ref{sec:exp_mixture}), and loss-based reweighting all beat their static counterparts on Mistral-7B, Llama-3.2-3B, and Qwen2.5-1.5B; we view this as an empirical regularity within the evaluated settings rather than a universal property across training stages and domains. \textbf{(T2) Offline and online methods are complementary, not strictly ordered:} in selection, offline NEAR/TSDS are competitive with online methods on the larger Mistral-7B but fall well behind on Llama-3.2-3B; in mixture, DoReMi's offline weights and ODM's online bandit each lead on different metrics and scales (ODM: best MMLU at 6B and best overall PPL at 30B; DoReMi: best overall PPL at 6B and best MMLU at 30B). A unified benchmark is required to make these complementarities visible. \textbf{(T3) Backbone sensitivity is comparable to algorithm-choice sensitivity:} the 5.8-to-13.4-point difference across backbones (F2) is of the same order as the spread between the best and worst dynamic methods on either backbone, so both \emph{whether} and \emph{which} dynamic method to use should be assessed per backbone. \textbf{(T4) Shared infrastructure is a benchmark primitive:} the 1--7\% equal-resource runtime reductions across LESS, TSDS, and MoE-SFT, as well as the 57\% wall-clock reduction of distributed LESS on 8$\times$H20, come from shared primitives (cached projections, ZeRO-3 gradient collection, reused embedding caches) rather than algorithmic changes, indicating that reproducible benchmarking and scalable deployment of data-centric methods are fundamentally the same engineering concern.

\section{Conclusion}
In this paper, we present \textsc{DataFlex}, a unified data-centric dynamic training framework built on top of LLaMA-Factory. \textsc{DataFlex} is designed to treat data as a first-class optimization object and provides a practical infrastructure for dynamically controlling how training data are selected, mixed, and weighted during large language model optimization. By unifying the three major paradigms of data-centric training---data selection, data mixture, and data reweighting---under a common framework, \textsc{DataFlex} reduces fragmentation in existing implementations and improves the reproducibility and comparability of data-centric research.

A key strength of \textsc{DataFlex} lies in its system design. Instead of introducing an isolated external pipeline, \textsc{DataFlex} replaces the training layer of LLaMA-Factory with extensible trainer abstractions and modular algorithm components, enabling seamless integration into existing training workflows. This design not only lowers the engineering barrier to adopting and comparing representative data-centric methods, but also provides a flexible foundation for future algorithmic extensions. In addition, by unifying shared model-dependent operations such as embedding extraction, model inference, and gradient computation, \textsc{DataFlex} makes it easier to implement dynamic training methods in a scalable and reusable manner.

Experimental results demonstrate that dynamic data-centric methods implemented in \textsc{DataFlex} outperform static full-data training on MMLU in the evaluated data-selection settings. For data mixture, DoReMi and ODM improve MMLU at both evaluated token budgets and exhibit complementary strengths across domains and metrics, while the unified implementations also achieve runtime improvements over the original codebases. We hope that \textsc{DataFlex} can serve as a useful infrastructure for reproducible research on data-centric large language model training and facilitate future studies on more adaptive, efficient, and principled data--model interaction.

\clearpage

\bibliographystyle{plainnat}
\bibliography{main}

\clearpage

\beginappendix

\section{Author Contributions}

\newcommand{\ProjectLeader}{\textcolor{red!70!black}{\textit{Project Leader}}}
\newcommand{\ProjectFounder}{\textcolor{blue!70!black}{\textit{Project Founder}}}
\newcommand{\CoreContributor}{\textcolor{green!50!black}{\textit{Core Contributor}}}
\newcommand{\Contributor}{\textcolor{green!80!black}{\textit{Contributor}}}
\newcommand{\ProjectSupervisor}{\textcolor{purple!70!black}{\textit{Project Supervisor}}}
\newcommand{\CorrespondingAuthor}{\textcolor{orange!80!black}{\textit{Corresponding Author}}}

\begin{itemize}
    \item Hao Liang: \ProjectLeader, \ProjectFounder, \CoreContributor; led the algorithmic and system design, extended data selection methods to large-scale settings, and contributed to manuscript preparation.
    
    \item Zhengyang Zhao: \ProjectFounder, \CoreContributor; design and implementation of data selection and reweighting interface.

    \item Mingrui Chen: \CoreContributor; design and implementation of data mixture interfaces. Enhancement of data mixture methods and execution of data mixture experiments.
    
    \item Meiyi Qiang: \CoreContributor; contributed to the NICE data selection algorithm, improved the LESS selector, and led selector and reweighting experiments and analysis.
    
    \item Lu Ma: \Contributor; implementation of the first LESS data selector.
    
    \item Rongyi Yu: \Contributor; implementation of offline data selection algorithms.
    
    \item Hengyi Feng: \Contributor; execution of selector experiments and manuscript revision and writing.
    
    \item Shixuan Sun: \Contributor; implementation of data mixture algorithms.
    
    \item Zimo Meng: \Contributor; contributions to the \textsc{DataFlex} system development.
    
    \item Xiaochen Ma: \Contributor; architectural discussions and design of the \textsc{DataFlex} system.
    
    \item Xuanlin Yang: \Contributor; implementation of the zero-order data selector.
    
    \item Qifeng Cai: \Contributor; manuscript revision and writing.
    
    \item Ruichuan An: \Contributor; manuscript revision and writing.
    
    \item Bohan Zeng: \Contributor; manuscript revision and writing.

    \item Zhen Hao Wong: \Contributor; manuscript revision and writing.
    
    \item Chengyu Shen: \Contributor; manuscript revision and writing.

    \item Runming He: \Contributor; manuscript revision and writing.

    \item Zhaoyang Han: \Contributor; manuscript revision and writing.
    
    \item Yaowei Zheng: \ProjectSupervisor; project supervision.
    
    \item Fangcheng Fu: \ProjectSupervisor; project supervision.

    \item Conghui He: \ProjectSupervisor; project supervision.
    
    \item Bin Cui: \ProjectSupervisor; project supervision.
    
    \item Zhiyu Li: \ProjectSupervisor; project supervision.
    
    \item Weinan E: \ProjectSupervisor; project supervision.
    
    \item Wentao Zhang: \CorrespondingAuthor, \ProjectSupervisor; manuscript writing and project supervision.
\end{itemize}

\section{Formal Definitions of the Three Paradigms}
\label{app:preliminaries}

This appendix expands the brief introduction of \textsc{DataFlex}'s three paradigms in Section~\ref{sec:framework} with formal definitions, a unified comparison, equivalences and reductions, and a discussion of online versus offline decisions.

\paragraph{Notation.}
Let $\mathcal{D} = \{(x_i, y_i)\}_{i=1}^{N}$ denote a training corpus of $N$ samples, partitioned into $K$ disjoint domains $\mathcal{D} = \bigcup_{k=1}^{K} \mathcal{D}_k$. Let $\theta$ denote model parameters and $\ell(\theta; x_i, y_i)$ the per-sample loss. The standard training objective is
\begin{equation}
\mathcal{L}(\theta) = \frac{1}{N} \sum_{i=1}^{N} \ell(\theta; x_i, y_i).
\end{equation}

\subsection{Data Selection}
Data selection seeks a high-value subset $\mathcal{S}^* \subseteq \mathcal{D}$ such that training on $\mathcal{S}^*$ yields performance comparable to or better than training on the full corpus, subject to a budget $B$:
\begin{equation}
\label{eq:selection}
\mathcal{S}^* = \arg\max_{\mathcal{S} \subseteq \mathcal{D},\; |\mathcal{S}| \leq B} \;\operatorname{Perf}\bigl(\theta^*(\mathcal{S})\bigr),
\end{equation}
where $\theta^*(\mathcal{S})$ denotes parameters obtained by training on $\mathcal{S}$ and $\operatorname{Perf}(\cdot)$ measures downstream performance. The decision is \emph{binary} at the \emph{sample level}, applied \emph{before} a sample enters the training batch.

\subsection{Data Mixture}
Data mixture optimizes the sampling probability distribution over domains, finding domain-level weights $\mathbf{w} \in \Delta^{K-1}$ (the probability simplex) that minimize a validation criterion:
\begin{equation}
\label{eq:mixture}
\mathbf{w}^* = \arg\min_{\mathbf{w} \in \Delta^{K-1}} \;\mathcal{L}_{\mathrm{val}}\bigl(\theta^*(\mathbf{w})\bigr),
\end{equation}
where $\theta^*(\mathbf{w})$ is trained by sampling from domain $\mathcal{D}_k$ with probability $w_k$. The decision is a \emph{continuous distribution} at the \emph{domain level}, applied \emph{during data sampling}.

\subsection{Data Reweighting}
Data reweighting assigns a continuous scalar weight $\lambda_i \geq 0$ to each training sample, modifying the training objective to
\begin{equation}
\label{eq:reweight}
\mathcal{L}_{\mathrm{reweight}}(\theta) = \sum_{i=1}^{N} \lambda_i \cdot \ell(\theta; x_i, y_i).
\end{equation}
Unlike selection, reweighting retains all samples but modulates their influence on gradient updates. The decision is a \emph{continuous scalar} at the \emph{sample level}, applied \emph{during backpropagation}.

\subsection{Comparison}

\begin{table}[h]
\centering
\caption{Comparison of the three data-centric paradigms unified by \textsc{DataFlex}. The three differ along control granularity, decision type, and the point at which they intervene in the training pipeline, motivating three corresponding trainer abstractions.}
\label{tab:paradigm_comparison}
\renewcommand{\arraystretch}{1.05}
\begin{tabular}{lccc}
\toprule
\textbf{Dimension}        & \textbf{Selection}                  & \textbf{Mixture}                & \textbf{Reweighting} \\
\midrule
Granularity               & Sample                              & Domain                          & Sample \\
Decision type             & Binary                              & Distribution                    & Continuous \\
Intervention point        & Before batch                        & During sampling                 & Backpropagation \\
Formal variable           & $\mathcal{S} \subseteq \mathcal{D}$ & $\mathbf{w} \in \Delta^{K-1}$   & $\lambda_i \in \mathbb{R}_{\geq 0}$ \\
\textsc{DataFlex} trainer & Select Trainer                      & Mix Trainer                     & Weight Trainer \\
\bottomrule
\end{tabular}
\end{table}

As summarized in Table~\ref{tab:paradigm_comparison}, selection and reweighting both operate at the sample level, but selection makes a discrete inclusion decision \emph{before} training on a batch, while reweighting assigns continuous importance scores that modulate gradient contributions \emph{during} backpropagation. Mixture, by contrast, operates at the coarser domain level and adjusts sampling probabilities \emph{during data sampling}. These distinctions motivate the three dedicated trainer abstractions in Section~\ref{sec:architecture}.

\subsection{Equivalences and Reductions}
\label{app:equivalences}

Although Equations~(\ref{eq:selection})--(\ref{eq:reweight}) treat the three paradigms separately, they can be viewed as instances of a more general per-sample reweighting formalism~\citep{adapt2026}:

\begin{itemize}
\item \textbf{Selection as 0/1 reweighting.} Setting $\lambda_i \in \{0, 1\}$ in Equation~(\ref{eq:reweight}) recovers data selection: samples with $\lambda_i = 0$ contribute nothing to the gradient and are effectively excluded.
\item \textbf{Mixture as domain-level reweighting.} If all samples in domain $\mathcal{D}_k$ share the same weight $\lambda_i = w_k / |\mathcal{D}_k|$ for $\mathbf{w} \in \Delta^{K-1}$, then Equation~(\ref{eq:reweight}) recovers the data-mixture objective.
\item \textbf{General reweighting.} Allowing $\lambda_i \in \mathbb{R}_{\geq 0}$ to vary freely across samples recovers the general reweighting objective.
\end{itemize}

This unification is conceptually useful but does not eliminate the engineering distinctions: selection and mixture make decisions \emph{before} or \emph{during} batch construction (and can therefore avoid forwarding excluded samples), while reweighting takes effect \emph{during} backpropagation (and therefore must process every sample at least up to its loss). \textsc{DataFlex} retains three separate trainer abstractions to expose these system-level differences while still supporting the unified view through a shared registry and component interface.

\subsection{Online vs.\ Offline Decisions}
\label{app:online_offline}

A second axis orthogonal to paradigm is the \emph{frequency} at which a decision is updated. Offline methods compute their decision once before training begins and use it throughout. Online methods update their decision based on the evolving model and optimizer state. \textsc{DataFlex} supports both regimes uniformly: offline methods register a one-shot \texttt{prepare} hook invoked before training, while online methods register a periodic \texttt{update} hook controlled by \texttt{warmup\_step}, \texttt{update\_step}, and \texttt{update\_times}. From the algorithm's perspective the only difference is which hook to implement; the trainer-side scheduling logic is shared.

\subsection{Relation to Prior Formalizations}
\label{app:prior_formalizations}

Several prior works adopt closely related but not identical formalizations. \citet{xia2024less} formalize selection as a per-example influence maximization problem with respect to a held-out validation gradient---a special case of Equation~(\ref{eq:selection}) with $\operatorname{Perf}$ approximated linearly. \citet{xie2023doremi} formalize mixture as a Group-DRO problem over excess losses, which solves a robust variant of Equation~(\ref{eq:mixture}). \citet{sow2025dynamic} formalize reweighting as a function of per-sample loss, instantiating Equation~(\ref{eq:reweight}) with $\lambda_i = f(\ell_i)$ for some monotone $f$. The objective views above are deliberately abstract: each individual algorithm corresponds to a specific instantiation of the relevant equation, and \textsc{DataFlex} provides the infrastructure required to execute these instantiations under modern distributed training.

\section{Extended Related Work}
\label{app:related}

This appendix expands the related work summarized in Section~\ref{sec:related} with detailed discussion of representative methods.

\subsection{Data Selection}

\paragraph{Offline Data Selection.}
Offline methods determine the selected subset prior to model training, typically based on signals such as quality, diversity, or difficulty~\citep{bai2024survey, JCST-2509-15948, liang2025dataflow}. A common line of work relies on language models or auxiliary scoring models to assess data quality. \citet{du2023mods} use DeBERTa~\citep{he2020deberta} to score data quality and combine it with k-center greedy selection to improve diversity. \citet{chen2023alpagasus} leverage ChatGPT to evaluate the quality of instruction data and retain high-quality instances. \citet{xu2023rethinking} use GPT-4 to rewrite training examples to increase complexity and then refine the dataset by reducing redundancy. \citet{liu2023makes} train scoring models from data labeled by gpt-3.5-turbo-0613 to estimate both quality and complexity, while \citet{lu2023instag} use GPT-4-generated tags to characterize instance complexity and diversity. \citet{parkar2024selectllm} combine clustering with GPT-4-based filtering to balance quality and coverage. Offline methods are easy to deploy and incur no training-time overhead, but cannot adapt to changing model states.

\paragraph{Online Data Selection.}
Online methods update the selection policy during training using current model feedback. LESS~\citep{xia2024less} estimates per-example influence on target objectives through gradient approximation, enabling efficient instruction tuning with subset training. LearnAlign~\citep{li2025learnalign} extends this idea to reinforcement learning by aligning data selection with policy gradient directions. NICE~\citep{wangnice} generalizes online selection to settings with non-differentiable evaluation metrics through black-box optimization. More generally, theoretical work such as Data Selection via Optimal Control~\citep{gu2024data} and Data Efficacy for Language Model Training~\citep{dai2025data} formalizes data selection as optimization over training trajectories. Online selection is more flexible and often more effective, but requires tighter integration with the training loop and efficient access to intermediate model signals.

\subsection{Data Mixture}

\paragraph{Offline Data Mixture.}
Offline mixture methods estimate domain weights before large-scale training. DoReMi~\citep{xie2023doremi} proposes a two-stage strategy in which a small proxy model is first trained and used with a Group-DRO-style objective to derive domain weights, after which the full model is trained on data resampled by these weights. DoGE~\citep{fan2023doge} formulates domain reweighting as a bilevel optimization problem and learns proportions robust to varying target mixtures. RegMix~\citep{liu2024regmix} estimates optimal domain combinations through proxy-based regression. These methods provide principled initial mixtures with limited runtime overhead, but their estimated weights remain fixed.

\paragraph{Online Data Mixture.}
Online mixture methods adjust domain proportions dynamically. ODM~\citep{albalak2023efficient} formulates domain allocation as a multi-armed bandit and updates ratios from online loss observations. Aioli~\citep{chen2024aioli} models cross-domain interactions by estimating how one domain influences validation loss of another. Adaptive Data Optimization~\citep{jiang2024adaptive} fits per-domain loss curves and reallocates training mass. Sheared LLaMA~\citep{xia2023sheared} adjusts per-domain weights using reference losses from the original model. MoE-SFT~\citep{zhu2024dynamicdatamixingmaximizes} targets mixture-of-experts models and updates domain proportions online from gate-load divergence across domains, providing an architecture-specific alternative to loss-based signals.

\subsection{Data Reweighting}

\paragraph{Classical Sample Reweighting.}
Prior to LLMs, sample reweighting was studied in robust and curriculum learning. \citet{ren2018learning} derive sample weights from gradient direction on a held-out validation set to downweight noisy or mislabeled examples. \citet{shu2019meta} propose Meta-Weight-Net, learning an explicit weighting function via meta-learning. These works establish algorithmic foundations but were designed for moderate-scale classification.

\paragraph{Reweighting in LLM Training.}
Adapting reweighting to large-scale LLM training raises new challenges as the cost of computing per-sample signals grows with model and corpus size. \citet{thakkar2023self} use self-influence guided reweighting based on training dynamics. \citet{chen2024hard} show that emphasizing hard samples improves continual training generalization. \citet{sow2025dynamic} introduce a dynamic loss-based strategy where higher-loss samples receive larger weights, demonstrating improved convergence and overall performance. \citet{adapt2026} propose ADAPT, computing per-sample weights from similarity between training sample embeddings and a validation anchor set, and provide a unified formalization showing that selection and mixing are special cases of reweighting with binary and domain-level weights, respectively.

\subsection{Data-Centric Training Systems}

Beyond algorithm-specific work, recent research has explored system support for data-centric foundation model training. Mixtera~\citep{bother2025mixtera} proposes a declarative data plane that allows users to specify which samples should be used, in what proportion, and in what order, while supporting dynamic mixture adjustment based on model feedback. In contrast, \textsc{DataFlex} unifies multiple data-centric paradigms---selection, mixture, and reweighting---within a single training framework, integrating pluggable algorithm components directly into the optimization loop on top of LLaMA-Factory rather than providing a standalone data access layer.

\section{Engineering Comparison with Original Implementations}
\label{app:comparison}

We compare \textsc{DataFlex} against the original codebases of two representative methods---LESS for data selection and DoReMi for data mixture---along three engineering dimensions: distributed training support, model and architecture compatibility, and workflow integration.

\begin{table}[h]
\centering
\caption{Engineering comparison between original implementations and \textsc{DataFlex}.}
\label{tab:eng_comparison}
\setlength{\tabcolsep}{6pt}
\renewcommand{\arraystretch}{1.05}
\begin{tabular}{lccc}
\toprule
\textbf{Dimension} & \textbf{LESS} & \textbf{DoReMi} & \textbf{\textsc{DataFlex}} \\
\midrule
Multi-GPU / multi-node       & \ding{56} & \ding{56} & \ding{52} \\
ZeRO-3 / FSDP support        & \ding{56} & \ding{56} & \ding{52} \\
Standard model interface     & \ding{56} & \ding{56} & \ding{52} \\
Up-to-date dependencies      & \ding{56} & ---    & \ding{52} \\
Unified config entry point   & \ding{56} & \ding{56} & \ding{52} \\
Config-based domain addition & ---    & \ding{56} & \ding{52} \\
\bottomrule
\end{tabular}
\end{table}

\subsection{Data Selection: LESS}

The original LESS codebase validates the algorithmic concept but exhibits several engineering limitations.

\paragraph{Distributed Training.}
The original LESS targets single-GPU execution. \textsc{DataFlex} integrates with DeepSpeed ZeRO and FSDP, reconstructing full-rank gradients and Adam states from sharded parameters, making gradient-based selection applicable to larger models and longer contexts.

\paragraph{Dependency Management.}
The original codebase pins specific dependency versions (e.g., \texttt{transformers 4.36.2}, \texttt{torch 2.1.2}); upgrades trigger compilation errors or API conflicts with newer models. \textsc{DataFlex} inherits LLaMA-Factory's actively maintained dependency chain, avoiding these issues.

\paragraph{Pipeline Integration.}
The original repository organizes its pipeline as standalone scripts for data preparation, gradient extraction, and training, requiring manual coordination. \textsc{DataFlex} consolidates these into a unified pipeline with a single YAML entry point.

\paragraph{Online Extension.}
The original LESS operates offline with a fixed reference model. \textsc{DataFlex} extends it to an online setting by recomputing projected gradients at configurable intervals using the evolving model and optimizer state, enabling LESS-style gradient-based selection to adapt to the model's learning dynamics---a capability absent in the original implementation.

\subsection{Data Mixture: DoReMi}

The official DoReMi implementation demonstrates the core algorithmic idea, but its engineering design presents barriers to scalable and flexible deployment. \textsc{DataFlex} re-implements DoReMi as a pluggable mixer component while preserving the original algorithmic logic.

\paragraph{Distributed Training.}
The official DoReMi codebase lacks native support for multi-node training and mature integration with frameworks such as ZeRO and FSDP. \textsc{DataFlex} provides full distributed training support, enabling stable execution across multi-node, multi-GPU infrastructures without additional engineering effort.

\paragraph{Model Interface Decoupling.}
The official implementation introduces structural dependencies between data preprocessing and the model interface, requiring the forward pass to return non-standard outputs such as per-token loss and reference loss. \textsc{DataFlex} decouples the mixer component from the model by relying solely on the standard causal language model interface, enabling seamless adaptation to evolving open-source models (e.g., Qwen, LLaMA) with minimal modification.

\paragraph{Data Pipeline.}
The official codebase requires strict offline data preprocessing and fragmented script modifications to introduce new datasets. \textsc{DataFlex} replaces this with a unified \texttt{mixture\_manager} and dynamically reconstructed distributed dataloaders, allowing new data domains to be added via configuration without offline preprocessing or code modifications.

\subsection{Summary}

Table~\ref{tab:eng_comparison} summarizes the engineering differences across all three dimensions. Both original codebases share similar limitations in distributed training and workflow integration, which \textsc{DataFlex} addresses through its unified architecture built on LLaMA-Factory.

\section{Detailed Experimental Setup}
\label{app:hparams}

\subsection{Data Selection and Reweighting}

\paragraph{Dataset.}
We use a 100k-example subset of Open-Hermes-2.5 as the training pool. The validation and test sets are constructed from the MMLU validation and test splits.

\paragraph{Models.}
Mistral-7B-v0.1~\citep{2023arXiv231006825J} and Llama-3.2-3B~\citep{grattafiori2024llama}, both initialized with their official pre-trained weights.

\paragraph{Algorithms.}
We compare four categories of methods:
(1) \textbf{Online selection:} LESS~\citep{xia2024less} (extended to online as described in Section~\ref{sec:algorithms}), NICE~\citep{wangnice}, Loss, Delta Loss~\citep{zhang2025preference};
(2) \textbf{Offline selection:} NEAR, TSDS~\citep{liu2024tsds};
(3) \textbf{Reweighting:} Reweight;
(4) \textbf{Baselines:} training on the complete 100k dataset (static full-data), and a Random-selection baseline that samples examples uniformly at a matched budget. The Random and full-data settings are used as reference points; they are not counted among the ten integrated data-centric algorithms.

\paragraph{Training.}
We adopt LoRA~\citep{hu2022lora} on all linear layers with $r{=}32$ and $\alpha{=}64$. All models are trained for 1 epoch with AdamW, cosine learning-rate schedule, warmup ratio 0.1, learning rate $5\times 10^{-7}$, global batch size 8 (per-device 1 with 8 gradient-accumulation steps). For online methods we use \texttt{warmup\_step}=100, \texttt{update\_step}=50, \texttt{update\_times}=30. For fairness, the full-data baseline uses the same warmup. All experiments run on 8$\times$NVIDIA H20 GPUs.

\paragraph{Larger backbones.}
For Qwen3-14B (Table~\ref{tab:selection_qwen14b}), we train on 8$\times$H20 GPUs with a learning rate of $10^{-5}$, a 50-step warmup, and a 50-step selection interval. For Llama-3.1-70B (Table~\ref{tab:selection_llama70b}), we run distributed online selection on 16$\times$H20 GPUs using a 10K-example subset of Open-Hermes-2.5, with a learning rate of $10^{-5}$, a 100-step warmup, and a 200-step selection interval. Checkpoints are chosen by loss-based early stopping (step 200 for Qwen3-14B and step 400 for Llama-3.1-70B); by these checkpoints, at least one full data-selection update has been completed. In both cases, the base model is evaluated at step 0.

\subsection{Data Mixture}

\paragraph{Dataset.}
SlimPajama~\citep{cerebras2023slimpajama}, a deduplicated English pretraining corpus derived from RedPajama~\citep{weber2024redpajama}, comprising seven domains: CommonCrawl (CC), C4, GitHub, Book, ArXiv, Wikipedia, StackExchange (SE). Two subsets at different scales: SlimPajama-6B and SlimPajama-30B. Both are randomly sampled to preserve natural domain proportions: CC 54.1\%, C4 28.7\%, GitHub 4.2\%, Book 3.7\%, ArXiv 3.4\%, Wikipedia 3.1\%, SE 2.8\%. These natural proportions serve as the default baseline mixture and as initial weights for dynamic methods.

\paragraph{Models.}
We use the Qwen2.5 family~\citep{qwen2.5}. The target model (DoReMi Step~3 and ODM) is Qwen2.5-1.5B. For DoReMi's reference and proxy models we use Qwen2.5-0.5B. For the additional backbone experiment on SlimPajama-6B (Table~\ref{tab:data_mixture_llama}), the target model is Llama-3.2-3B~\citep{grattafiori2024llama} and DoReMi's reference and proxy models are Llama-3.2-1B. All models are trained from scratch with random initialization to isolate the effect of mixture strategies.

\paragraph{Baseline.}
Qwen2.5-1.5B trained from scratch with a static mixer at default proportions. For 6B: peak LR $5\times 10^{-5}$, per-device batch 8. For 30B: peak LR $2.5\times 10^{-4}$, per-device batch 16.

\paragraph{DoReMi.}
The three-step procedure is as follows.
\begin{itemize}
\item \textbf{Step~1:} Reference model (Qwen2.5-0.5B) trained from scratch with default proportions.
\item \textbf{Step~2:} Proxy model (Qwen2.5-0.5B) trained with the DoReMi algorithm: per-domain excess losses relative to the Step~1 reference are computed and domain weights updated via exponentiated gradient ascent. Weights are initialized uniformly ($\alpha_i^{(0)} = 1/K$, $K{=}7$), with step size $\eta = 0.1$ and smoothing $\varepsilon = 0.01$.
\item \textbf{Step~3:} Target model (Qwen2.5-1.5B) trained from scratch using the optimized weights from Step~2 as a static mixture.
\end{itemize}
For 6B: LR $5\times 10^{-5}$, per-device batch 8 across all steps (Step~2 uses gradient accumulation 8); weights updated every 100 steps after a 1{,}000-step warmup. For 30B: Steps~1--2 use LR $2.5\times 10^{-4}$, batch 16; weights updated every 500 steps after a 2{,}000-step warmup; Step~3 uses LR $5\times 10^{-4}$. Step~2 performs 100 weight updates in both settings.

\paragraph{ODM.}
ODM~\citep{albalak2023efficient} dynamically adjusts domain weights during a single training pass using Exp3 with exponential-weight moving average. Qwen2.5-1.5B is trained from scratch with weights initialized to default proportions. After a 2{,}000-step warmup (initial proportions held fixed), weights are updated periodically based on observed per-domain batch losses. Both settings share $\alpha = 0.90$ and reward scale 15. For 6B: LR $5\times 10^{-5}$, batch 8, $\varepsilon_{\min}=0.01$, clipping $-10$, weight updates every 500 steps. For 30B: LR $1.5\times 10^{-4}$, batch 16, $\varepsilon_{\min}=0.03$, clipping $-5$, weight updates every 1{,}000 steps.

\paragraph{Common.}
All experiments train for 1 full epoch with linear LR decay and 5\% warmup ratio. We use BFloat16 mixed precision, DeepSpeed ZeRO Stage-3~\citep{rasley2020deepspeed}, FlashAttention-2~\citep{dao2023flashattention}, and the Qwen tokenizer. Random seed 42. The 6B experiments run on a single node with 8 H20 GPUs; the 30B experiments scale to 4 nodes with 8 H20 GPUs each (32 GPUs total) coordinated via \texttt{torchrun}.

\subsection{System Efficiency}

\paragraph{LESS efficiency comparison.}
We use \texttt{Llama-2-7b-hf}~\citep{touvron2023llama} as the base model. For LESS we adopt its default configurations; for \textsc{DataFlex} we use the settings in the data selection section above. To evaluate single-cycle throughput we adjust to \texttt{warmup\_step}=100, \texttt{update\_step}=100, \texttt{update\_times}=1 on a single H20 GPU. We additionally evaluate parallel scalability on 8$\times$H20 GPUs with \texttt{warmup\_step}=100, \texttt{update\_step}=200, \texttt{update\_times}=3.

\paragraph{TSDS efficiency comparison (Appendix~\ref{app:tsds}).}
\label{app:tsds}
For each tokenized sample, \textsc{DataFlex} encodes it into a sentence-level embedding using \texttt{qwen3-embed-0.6B}, then performs nearest-neighbor retrieval and KDE-based density estimation. The embedding stage supports both Sentence-Transformer-based backends and vLLM-based inference. TSDS hyperparameters are fixed: \texttt{max\_K}=5000 (similarity-search neighbors), \texttt{kde\_K}=1000 (KDE neighbors), \texttt{sigma}=0.75, \texttt{alpha}=0.6, \texttt{C}=5. We compare the original TSDS operator and \textsc{DataFlex} under two scaling settings: (i) varying training set size from 5k to 100k while fixing validation at 1k queries, and (ii) varying validation set size from 50 to 1k while fixing training at 10k examples.

\paragraph{MoE-SFT online mixer (Appendix~\ref{app:moe_sft}).}
We train LLaMA-MoE-v1-3.5B-2/8~\citep{zhu2024llamamoebuildingmixtureofexpertsllama} with full-parameter SFT for 2{,}000 steps on four instruction-tuning domains (code, math, ShareGPT, Orca) under DeepSpeed ZeRO-3 with BF16, LR $2\times 10^{-5}$ (cosine schedule, warmup ratio 0.03), max sequence length 2{,}048. The mixer uses $\eta = 10.0$, $c = 0.05$, update interval $m = 100$, and 60 warmup steps with uniform sampling. Following MoE-SFT's protocol we evaluate knowledge \& reasoning via MMLU~\citep{hendryckstest2021} (5-shot), BBH~\citep{suzgun2022challenging} (3-shot), QA (ARC-e/c~\citep{clark2018thinksolvedquestionanswering}, BoolQ~\citep{clark2019boolqexploringsurprisingdifficulty}), and code generation via MBPP~\citep{austin2021programsynthesislargelanguage} (zero-shot, Pass@1).

\section{Extended Discussion of Empirical Results}
\label{app:discussion}

\subsection{Additional Selection and Reweighting Results}
\label{app:selection_additional}

To investigate whether the effectiveness of selection and reweighting methods is sensitive to model scale or confined to particular benchmarks or tasks, we further conduct independent runs on Llama-3.2-1B and Llama-3.2-3B for within-family scale comparison, and on Llama-3.2-3B and Mistral-7B-v0.1 across multiple benchmarks.

\paragraph{Within-family scale comparison.}
Table~\ref{tab:selection_within_family} compares Llama-3.2-1B and Llama-3.2-3B under the same data split, LoRA configuration, and learning rate. 
The best method differs by benchmark and scale, showing that the effect of data-centric training is sensitive to the backbone and evaluation target rather than attributable to a model-size-only effect.

\begin{table}[t]
\centering
\small
\caption{Controlled within-family comparison on Llama-3.2-1B and Llama-3.2-3B at step 400. Both use the same data split, LoRA configuration ($r=32$, $\alpha=64$), and learning rate ($10^{-5}$). Scores are accuracy (\%).}
\label{tab:selection_within_family}
\setlength{\tabcolsep}{4pt}
\renewcommand{\arraystretch}{1.04}
\begin{tabular}{llcccc}
\toprule
\textbf{Backbone} & \textbf{Method} & \textbf{ARC-C} & \textbf{ARC-E} & \textbf{GSM8K} & \textbf{IFEval} \\
\midrule
\multirow{4}{*}{Llama-3.2-1B}
& Static     & \textbf{36.86} & 60.69 & 2.12 & 9.43 \\
& Delta Loss & 36.43 & 59.26 & 2.43 & 7.58 \\
& Reweight   & \textbf{36.86} & 61.36 & \textbf{2.50} & 7.95 \\
& TSDS       & 36.43 & \textbf{62.25} & 1.59 & \textbf{10.17} \\
\midrule
\multirow{4}{*}{Llama-3.2-3B}
& Static     & 46.50 & 72.77 & 3.49 & 10.17 \\
& Delta Loss & 46.50 & \textbf{73.53} & 3.26 & 9.43 \\
& Reweight   & 46.42 & 72.90 & \textbf{4.09} & 10.17 \\
& TSDS       & \textbf{47.10} & 73.11 & 2.73 & \textbf{12.38} \\
\bottomrule
\end{tabular}
\end{table}

\paragraph{Cross-benchmark evaluation.}
Tables~\ref{tab:selection_benchmarks_llama} and~\ref{tab:selection_benchmarks_mistral} extend the original MMLU evaluation to mathematical reasoning, general reasoning, instruction following, and grade-school mathematics. Different strategies lead on different benchmark--backbone pairs, so these results are reported as comparative evidence within the evaluated settings rather than as a universal method ranking.

\begin{table}[t]
\centering
\small
\caption{Cross-benchmark evaluation on Llama-3.2-3B at step 400. Scores are accuracy (\%).}
\label{tab:selection_benchmarks_llama}
\setlength{\tabcolsep}{4pt}
\renewcommand{\arraystretch}{1.03}
\begin{tabular}{lccccc}
\toprule
\textbf{Method} & \textbf{MATH-500} & \textbf{BBH} & \textbf{ARC-C} & \textbf{ARC-E} & \textbf{IFEval} \\
\midrule
Base       & 6.80 & 46.92 & 45.82 & 71.63 & 8.50 \\
Random     & 6.80 & 47.17 & 46.25 & 72.94 & 10.35 \\
Static     & 5.60 & 47.20 & 46.50 & 73.27 & 9.61 \\
Loss       & 6.80 & 47.06 & 46.50 & 72.52 & 10.17 \\
Delta Loss & 7.60 & 46.91 & 46.50 & 73.53 & 9.43 \\
Reweight   & 7.40 & 47.06 & 46.42 & 72.90 & 10.17 \\
NEAR       & \textbf{10.80} & 45.58 & 45.73 & 73.57 & 8.32 \\
TSDS       & 8.20 & \textbf{47.55} & 47.10 & 73.11 & \textbf{12.38} \\
LESS       & 7.60 & 47.41 & \textbf{47.44} & \textbf{74.49} & 8.32 \\
\bottomrule
\end{tabular}
\end{table}

\begin{table}[t]
\centering
\small
\caption{Cross-benchmark evaluation on Mistral-7B-v0.1 at step 800. Scores are accuracy (\%).}
\label{tab:selection_benchmarks_mistral}
\setlength{\tabcolsep}{3.6pt}
\renewcommand{\arraystretch}{1.03}
\begin{tabular}{lcccccc}
\toprule
\textbf{Method} & \textbf{MATH-500} & \textbf{BBH} & \textbf{ARC-C} & \textbf{ARC-E} & \textbf{GSM8K} & \textbf{IFEval} \\
\midrule
Base       & 12.80 & 56.14 & 54.01 & 79.59 & 11.75 & 17.01 \\
Random     & \textbf{15.80} & 58.72 & 54.95 & 80.26 & 8.79 & 20.70 \\
Static     & 13.80 & 57.21 & 54.10 & 80.01 & 11.68 & 19.04 \\
Loss       & \textbf{15.80} & \textbf{59.07} & 54.44 & 79.67 & 8.11 & \textbf{23.48} \\
Delta Loss & 14.60 & 58.85 & 54.52 & 78.58 & 10.39 & 23.29 \\
Reweight   & 14.20 & 58.78 & 55.63 & 80.56 & 9.17 & 21.44 \\
NEAR       & 12.80 & 55.95 & 52.13 & 74.37 & \textbf{27.29} & 18.67 \\
TSDS       & 14.40 & 58.27 & 56.23 & 79.50 & 8.49 & 19.78 \\
LESS       & 12.20 & 58.38 & \textbf{57.76} & \textbf{81.14} & 12.81 & 20.70 \\
\bottomrule
\end{tabular}
\end{table}

\subsection{Data Selection and Reweighting (Figure~\ref{fig:two-models})}

\paragraph{Per-method analysis on Mistral-7B.}
LESS achieves the best final accuracy of 0.452, outperforming the static baseline (0.394) by 5.8 points. Reweight and TSDS tie at 0.429, followed by NEAR (0.419), NICE (0.418), and Delta Loss (0.412). The Loss-only selector (0.400) and Random (0.393) perform comparably to the static baseline, suggesting that simple loss-based or random subsampling provides limited benefits for this larger model. Notably, the offline methods (NEAR, TSDS) converge faster in early training (steps 200--600) than most online methods, likely because their precomputed selections are already aligned with high-value samples from the start.

\paragraph{Per-method analysis on Llama-3.2-3B.}
On the smaller backbone, the gap between dynamic methods and the static baseline is even more pronounced. The static baseline plateaus at 0.319, while all online methods exceed 0.427. Reweight achieves the best final accuracy (0.453), closely followed by LESS (0.450) and Delta Loss (0.434). NICE reaches 0.428, Random reaches 0.431, and Loss attains 0.429---all substantially outperforming the static baseline. The offline methods (NEAR 0.344, TSDS 0.345) perform notably worse than online methods, indicating that dynamic, model-aware selection is more critical on this backbone.

\paragraph{Cross-backbone insights.}
Comparing across backbones reveals three patterns. First, online methods that leverage model-dependent signals---primarily LESS and Reweight, joined by Delta Loss on Llama-3.2-3B---consistently rank at the top, suggesting that exploiting the evolving model state is a particularly effective curation strategy in these settings. Second, the relative benefit of dynamic selection is backbone dependent: on Mistral-7B the best method (LESS) improves over the static baseline by 5.8 points, whereas on Llama-3.2-3B the best method (Reweight) achieves a 13.4-point gain. Because the two backbones differ in both size and model family, this difference should not be attributed to model capacity alone; the within-family comparison in Appendix~\ref{app:selection_additional} provides a controlled comparison across scales. Third, offline methods exhibit a notable trade-off: they provide a lower-overhead alternative but cannot adapt to the evolving model state, which limits their final accuracy on the smaller, faster-shifting Llama-3.2-3B where they trail every online method by roughly 8 points or more.

\paragraph{Practical guidance.}
Gradient-based methods (LESS) offer the highest final accuracy but incur the cost of periodic gradient collection and projection. Loss-based reweighting achieves competitive accuracy with lower overhead since it operates on readily available training losses. Offline methods provide a lightweight alternative when training-time computation is constrained. \textsc{DataFlex} enables all these trade-offs to be explored under identical experimental conditions.

\subsection{Data Mixture (Table~\ref{tab:data_mixture})}

\paragraph{SlimPajama-6B.}
On the 6B-token subset, both dynamic methods outperform the static baseline on MMLU, with ODM achieving the highest accuracy (26.04\%), followed by DoReMi (25.84\%) and the baseline (25.27\%). DoReMi achieves the best overall log perplexity (4.134). Per-domain, DoReMi achieves the lowest perplexity on CommonCrawl (4.108) and C4 (4.358)---the two largest domains---likely because its three-step optimization explicitly targets excess loss across all domains relative to the reference model. ODM yields the best perplexity on StackExchange (3.243), ArXiv (2.904), and Book (4.613)---smaller, more specialized domains---suggesting that its online Exp3 exploration effectively identifies and upweights underrepresented domains during training.

\paragraph{SlimPajama-30B.}
At 30B, both dynamic methods outperform the baseline on both MMLU and overall log perplexity. DoReMi achieves the highest MMLU (25.97\%), followed by ODM (25.63\%) and the baseline (25.51\%). ODM produces the best overall log perplexity (3.429), substantially outperforming DoReMi (3.562) and the baseline (3.584). Per-domain, ODM achieves the lowest perplexity on 6 of 7 domains. DoReMi remains best on C4 (3.503). At the 30B scale, the static baseline no longer wins any domain, indicating that dynamic mixing remains advantageous at the larger token budget.

\paragraph{Complementary characteristics.}
DoReMi, through its minimax-based excess-loss optimization, tends to improve perplexity on high-resource domains where most training data resides. ODM, through online bandit exploration, more aggressively explores and upweights minority domains, yielding stronger per-domain perplexity on specialized text types such as StackExchange, ArXiv, and Book. At 30B, the complementarity shifts from domains to metrics: ODM's exploration has more steps to converge and yields the best log perplexity on most domains, while DoReMi's optimized weights yield higher MMLU accuracy.

\paragraph{Computational trade-off.}
DoReMi (offline) requires training separate reference and proxy models before the main run but incurs no overhead during the target model's training, attractive when upfront cost is acceptable. ODM (online) avoids the multi-stage pipeline at the cost of lightweight per-step bookkeeping. \textsc{DataFlex} supports both under a unified interface, allowing practitioners to choose based on their computational budget.

Table~\ref{tab:mixture_gpu_hours} quantifies this trade-off on SlimPajama-30B. The target-model runs are closely aligned: DoReMi's Step~3 (1{,}001.92 GPU-hours) is comparable to the static baseline (1{,}008.96), and ODM's online bookkeeping adds about 2.3\% (1{,}031.68). Counting its reference and proxy stages, however, DoReMi's complete pipeline requires 1{,}806.08 GPU-hours, 1.79$\times$ that of the static baseline. The results in Table~\ref{tab:data_mixture} therefore compare methods under matched target-model training budgets, while the end-to-end cost characterizes the practical effectiveness--efficiency trade-off.

\begin{table}[t]
\centering
\small
\caption{End-to-end GPU-hours for data mixture on SlimPajama-30B, measured on the same 32$\times$H20 setup. The target-model runs of the three methods have closely aligned costs, while DoReMi's two auxiliary stages raise its complete pipeline to 1.79$\times$ the GPU-hours of the static baseline.}
\label{tab:mixture_gpu_hours}
\setlength{\tabcolsep}{6pt}
\renewcommand{\arraystretch}{1.04}
\begin{tabular}{llr}
\toprule
\textbf{Method / stage} & \textbf{Model} & \textbf{GPU-hours} \\
\midrule
Static baseline & Qwen2.5-1.5B & 1{,}008.96 \\
ODM & Qwen2.5-1.5B & 1{,}031.68 \\
\midrule
DoReMi Step~1 (reference) & Qwen2.5-0.5B & 442.88 \\
DoReMi Step~2 (proxy) & Qwen2.5-0.5B & 361.28 \\
DoReMi Step~3 (target) & Qwen2.5-1.5B & 1{,}001.92 \\
\textbf{DoReMi total} & -- & \textbf{1{,}806.08} \\
\bottomrule
\end{tabular}
\end{table}

\section{Online Data Mixer for MoE-SFT}
\label{app:moe_sft}

This appendix provides additional details for the MoE-SFT efficiency comparison reported in Section~\ref{sec:exp_mixture_efficiency}. The experiment evaluates \textsc{DataFlex}'s online data mixer on supervised fine-tuning of mixture-of-experts (MoE) language models, following the MoE-SFT setting~\citep{zhu2024dynamicdatamixingmaximizes}. The same experiment also demonstrates that the unified mixer interface generalizes to MoE-specific signals beyond loss-based feedback.

\subsection{Algorithm}

At every $m$-th training step, the mixer collects L1-normalized gate load vectors $\hat{O}_i \in \mathbb{R}^{E}$ from the last MoE layer for each domain $i$, where $E$ is the number of experts. It computes the mean pairwise L2 divergence
\begin{equation}
\Delta_i = \frac{1}{|D|}\sum_{j} \|\hat{O}_i - \hat{O}_j\|_2,
\end{equation}
and updates the sampling weights via
\begin{equation}
w_i^{(t)} \;\propto\; w_i^{(t-1)} \cdot \exp\!\bigl(\eta \cdot \Delta_i\bigr),
\end{equation}
followed by smoothing $w'_i = (1 - c)\,w_i + c / |D|$ and re-normalization. This upweights domains whose expert utilization diverges most from the ensemble. Gate parameters are frozen to ensure stable load signals.

\subsection{Setup}

We train LLaMA-MoE-v1-3.5B-2/8~\citep{zhu2024llamamoebuildingmixtureofexpertsllama} with full-parameter SFT for 2{,}000 steps on four instruction-tuning domains (code, math, ShareGPT, Orca) under DeepSpeed ZeRO-3 with BF16, learning rate $2\times 10^{-5}$ (cosine schedule, warmup ratio 0.03), and max sequence length 2{,}048. The mixer uses $\eta = 10.0$, $c = 0.05$, update interval $m = 100$, and 60 warmup steps with uniform sampling.

Following MoE-SFT's evaluation protocol, we assess knowledge \& reasoning via MMLU~\citep{hendryckstest2021} (5-shot), BBH~\citep{suzgun2022challenging} (3-shot), QA (averaged over ARC-e, ARC-c~\citep{clark2018thinksolvedquestionanswering}, BoolQ~\citep{clark2019boolqexploringsurprisingdifficulty}), and code generation via MBPP~\citep{austin2021programsynthesislargelanguage} (zero-shot, Pass@1). The aggregated efficiency and accuracy comparison is reported as Table~\ref{tab:online_data_mixer_comparison} in the main paper.

\subsection{Why MoE-SFT Is Used as the Reference for Data Mixture Efficiency}

Among the data-mixture algorithms integrated into \textsc{DataFlex}, MoE-SFT is the only method whose original codebase we were able to reproduce reliably at our experimental scale. The official DoReMi implementation pins outdated dependencies that are incompatible with current Llama and Qwen2.5 models, and the original ODM implementation lacks support for distributed training under DeepSpeed ZeRO-3 and assumes single-GPU operation; in both cases, retrofitting these constraints would require substantial reimplementation that is itself part of \textsc{DataFlex}'s contribution. MoE-SFT, by contrast, releases a maintained reference codebase that runs end-to-end under modern dependencies, providing the most faithful third-party comparison for the Data Mixture module.

\section{Additional Details for TSDS Efficiency}

This appendix provides additional context for the TSDS offline-selection efficiency comparison reported in Section~\ref{sec:exp_selection_efficiency}. The plotted runtimes (Figure~\ref{fig:tsds_efficiency}) are obtained on the same hardware as the LESS efficiency study, using the offline TSDS operator that scores samples by similarity to a validation anchor set in an embedding space.

\subsection{Setup}

The original TSDS implementation builds a similarity matrix between the training pool and the validation set, applies softmax over similarities, and selects the top-$k$ samples. \textsc{DataFlex}'s implementation reproduces this scoring procedure exactly but reuses several primitives provided by the framework: (i) a shared embedding cache that persists across multiple TSDS invocations within the same training run; (ii) blocked similarity computation that bounds peak memory and exploits BLAS-friendly tensor shapes; and (iii) deterministic top-$k$ selection compatible with the standard \textsc{DataFlex} dataloader rebuild path. The overall scoring algorithm is unchanged.

\subsection{Why TSDS Exhibits Smaller Speedups Than LESS}

The 1--5\% speedups for TSDS are generally smaller than the 3--7\% speedups observed for LESS in the same hardware setting. Two factors explain this gap. First, TSDS is dominated by dense matrix operations whose runtime is largely set by the underlying BLAS implementation; the room for system-level optimization is therefore limited. Second, TSDS is offline and runs once per experiment, whereas LESS runs repeatedly throughout training and benefits more from cached gradient projections and overlapped dataloader rebuilds. The fact that TSDS still shows a consistent improvement across all tested scales is therefore notable: it demonstrates that the framework's shared infrastructure does not impose overhead on offline workloads while still delivering modest gains.

\subsection{Scaling Regimes}

The two regimes plotted in Figure~\ref{fig:tsds_efficiency} are the most common in practice. Fixing the validation set and scaling the training pool (left panel) corresponds to selecting from a fresh, larger corpus; fixing the training pool and scaling the validation set (right panel) corresponds to refining the anchor set during iterative experiments. Both regimes exhibit linear growth, as expected from TSDS's pairwise similarity computation, with \textsc{DataFlex} consistently faster.

\section{Limitations and Broader Impact}
\label{app:limitations}

\subsection{Limitations}

\textsc{DataFlex} is a training framework, and its scope is intentionally bounded by what a single unified system can responsibly support and evaluate. We discuss the corresponding boundaries below, together with how the framework is designed to grow beyond them.

\paragraph{Scope of integrated algorithms.}
The current release integrates ten representative algorithms spanning the three paradigms studied in the paper---six data selection methods, three data mixture methods, and one data reweighting method. The distribution across paradigms mirrors the current state of the literature: data selection has accumulated a large and active body of algorithms, while data mixture and especially data reweighting remain comparatively smaller research areas with fewer widely cited methods at the LLM scale. Accordingly, our coverage is chosen to include the most widely cited and reproducibly usable methods in each paradigm, and the registry-based component interface is designed so that newly proposed methods---in any of the three paradigms---can be contributed upstream as the literature evolves.

\paragraph{Evaluation protocol.}
Our primary experiments follow the evaluation protocols established by the methods we compare against: instruction tuning with MMLU (as in LESS and related work) and pretraining on SlimPajama with MMLU and per-domain perplexity (as in DoReMi and ODM); additional selection experiments extend the evaluation to mathematical reasoning, general reasoning, and instruction following (Appendix~\ref{app:selection_additional}). This preserves apples-to-apples comparison with prior literature. Broader evaluation along additional axes---e.g., long-form generation, multilingual settings, and vision-language tasks---is orthogonal to \textsc{DataFlex}'s system design and can be conducted within the same framework using the evaluation hooks already exposed.

\paragraph{Scale of empirical study.}
The full set of integrated methods is evaluated on 3B--7B backbones for instruction tuning and on 1.5B--3B models pretrained with up to 30B tokens, which are representative of the regimes at which published data-centric methods are typically evaluated; larger backbones (Qwen3-14B and Llama-3.1-70B), as well as Llama-3.2-1B in the within-family comparison, are evaluated with a representative subset of methods. Evaluating all ten integrated algorithms at these larger scales, or pretraining with substantially larger token budgets, is primarily bounded by available compute: the 30B-token pretraining run in this paper already requires 32 H20 GPUs across four nodes. \textsc{DataFlex}'s system layer---DeepSpeed ZeRO-3, multi-node distributed execution, and shared full-rank gradient acquisition---is explicitly designed for such regimes, so that extending the empirical study is an engineering and compute investment rather than a methodological one.

\paragraph{Single-run results.}
Due to the high computational cost of the reported experiments (e.g., 32 H20 GPUs for SlimPajama-30B pretraining), results are based on single runs with a fixed seed, and we do not report error bars. Small differences between methods should therefore be interpreted as preliminary; multi-seed evaluation is left for future work.

\subsection{Broader Impact}

\textsc{DataFlex} is a general-purpose training framework whose contributions are at the system and infrastructure level rather than at the level of new model capabilities. Its societal implications therefore align with those of standard open-source LLM training infrastructure.

\paragraph{Positive impact.}
By unifying data selection, mixture, and reweighting under a single reproducible interface, \textsc{DataFlex} lowers the engineering barrier to studying data-centric training. This supports more transparent and comparable research on training-data efficiency, which is directly relevant to reducing the computational and environmental cost of LLM development, and broadens access to state-of-the-art data-centric methods for academic groups and smaller organizations with limited engineering resources.

\paragraph{Potential risks and mitigations.}
\textsc{DataFlex} does not release any pretrained model weights, scraped corpora, or application-specific artifacts; the methods it implements operate orthogonally to model capabilities and generated content. As with any general training framework, it could in principle be used to train models for harmful applications. We therefore encourage users to follow standard responsible-AI practices---careful data curation, evaluation for safety and bias, and consideration of downstream use---when deploying models trained with \textsc{DataFlex}.

\paragraph{Data and code release.}
We release \textsc{DataFlex} as an open-source framework with documentation, example configurations, and integration tests, so that independent researchers can verify and extend our results. The code is publicly available at \url{https://github.com/OpenDCAI/DataFlex}. All datasets used in this paper are publicly available (Open-Hermes-2.5, SlimPajama, MMLU); we provide preprocessing scripts to support reproducibility.

\end{document}